\documentclass[10pt,journal,compsoc]{IEEEtran}

\usepackage{epsfig}
\usepackage{graphicx}
\usepackage{amsmath}
\usepackage{amssymb}
\usepackage{enumitem}
\usepackage{bm}
\usepackage{booktabs}
\usepackage{subcaption}
\usepackage{multirow}
\DeclareMathOperator*{\argmax}{arg\,max}

\DeclareMathOperator*{\topk}{top}

\newcommand{\etal}{\textit{et al}. }
\newcommand{\ie}{\textit{i}.\textit{e}.}
\newcommand{\eg}{\textit{e}.\textit{g}.}

\makeatletter
\newcommand*\bigcdot{\mathpalette\bigcdot@{.7}}
\newcommand*\bigcdot@[2]{\mathbin{\vcenter{\hbox{\scalebox{#2}{$\m@th#1\bullet$}}}}}
\makeatother

\usepackage{mathtools}

\DeclarePairedDelimiter\floor{\lfloor}{\rfloor}

\usepackage{tabularx} 
\usepackage{adjustbox}

\ifCLASSOPTIONcompsoc
  \usepackage[nocompress]{cite}
\else
  \usepackage{cite}
\fi

\begin{document}
\title{Learning to Structure an Image\\ with Few Colors and Beyond}


\author{Yunzhong Hou, 
        Liang Zheng$^\star$, 
        and Stephen Gould
\IEEEcompsocitemizethanks{
\IEEEcompsocthanksitem $^\star$ Corresponding author.
\IEEEcompsocthanksitem Yunzhong Hou, Liang Zheng, and Stephen Gould are with Research School of Computer Science, Australian National University, Canberra, ACT 2601, Australia.
E-mail: \{firstname.lastname\}@anu.edu.au
\IEEEcompsocthanksitem Yunzhong Hou, Liang Zheng, and Stephen Gould are with Australian Centre for Robotic Vision.
}
\thanks{Manuscript received September 11, 2020; modified August 12, 2021; resubmitted-as-new January 15, 2022.}
\thanks{This work was supported by the ARC Discovery Early Career Researcher Award (DE200101283) and the ARC Discovery Project (DP210102801).}
}

\IEEEtitleabstractindextext{%
\begin{abstract}
Color and structure are the two pillars that combine to give an image its meaning. Interested in critical structures for neural network recognition, we isolate the influence of colors by limiting the color space to just a few bits, and find structures that enable network recognition under such constraints. 
To this end, we propose a color quantization network, ColorCNN, which learns to structure an image in limited color spaces by minimizing the classification loss. 
Building upon the architecture and insights of ColorCNN, we introduce ColorCNN+, which supports multiple color space size configurations, and addresses the previous issues of poor recognition accuracy and undesirable visual fidelity under large color spaces. Via a novel imitation learning approach, ColorCNN+ learns to cluster colors like traditional color quantization methods. This reduces overfitting and helps both visual fidelity and recognition accuracy under large color spaces. Experiments verify that ColorCNN+ achieves very competitive results under most circumstances, preserving both key structures for network recognition and visual fidelity with accurate colors. 
We further discuss differences between key structures and accurate colors, and their specific contributions to network recognition.
For potential applications, we show that ColorCNNs can be used as image compression methods for network recognition.

\end{abstract}

}

\maketitle
\IEEEdisplaynontitleabstractindextext
\IEEEpeerreviewmaketitle

\section{Introduction}\label{sec:Introduction}
\IEEEPARstart{I}{mages} use both color and structure to convey information. 
Curious about the critical structures for neural network recognition, we isolate the contribution of colors to the recognition task, and find key structures that alone enable network recognition under very small color spaces. 
As a combination of shapes, textures, and other visual cues, the underlying structure emerges from the arrangement of different colors. Particularly, structures are only well presented when there exists a sufficient set of colors. As such, we investigate the interplay between colors and structures to help our final goal. 

\begin{figure}
    \begin{subfigure}[b]{0.24\linewidth}
    \centering
        \includegraphics[width=\textwidth]{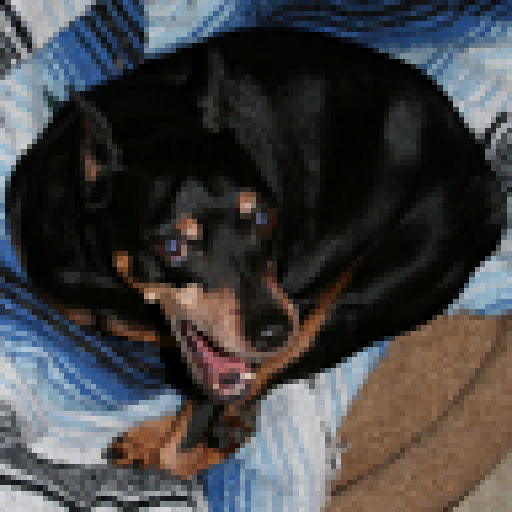}
    \end{subfigure}
    \hfill
    \begin{subfigure}[b]{0.24\linewidth}
    \centering
        \includegraphics[width=\textwidth]{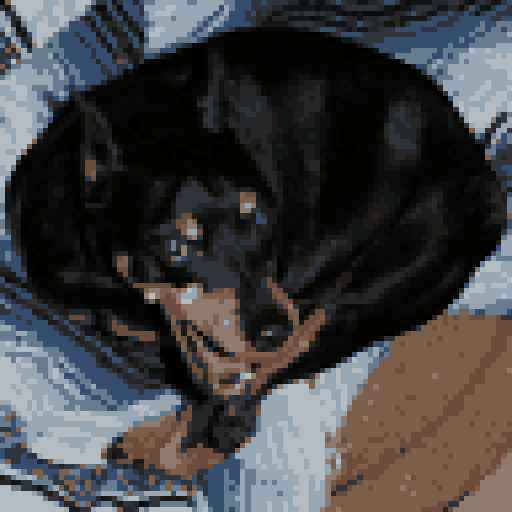}
    \end{subfigure}
    \hfill
    \begin{subfigure}[b]{0.24\linewidth}
    \centering
        \includegraphics[width=\textwidth]{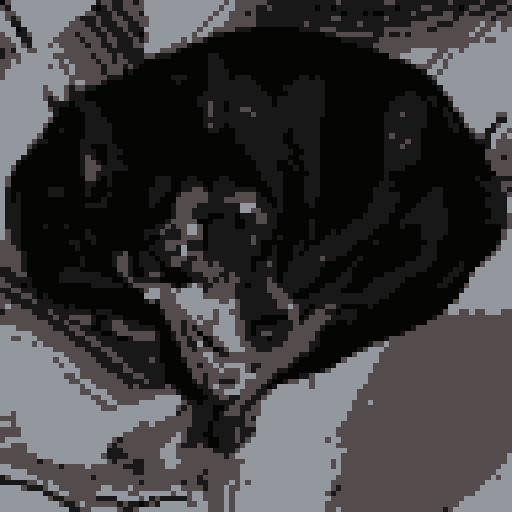}
    \end{subfigure}
    \hfill
    \begin{subfigure}[b]{0.24\linewidth}
    \centering
        \includegraphics[width=\textwidth]{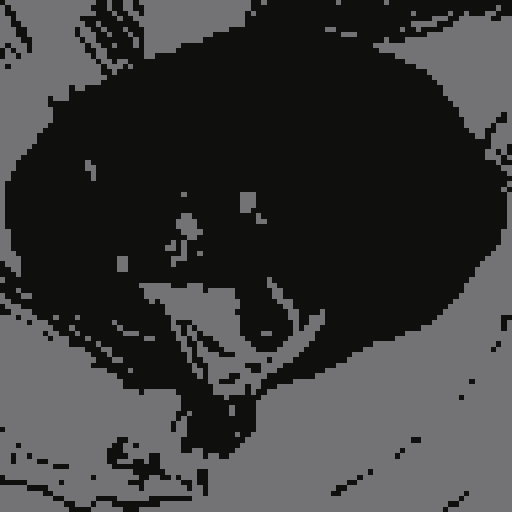}
    \end{subfigure}

    \begin{subfigure}[b]{0.24\linewidth}
    \centering
        \includegraphics[width=\textwidth]{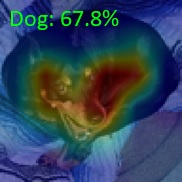}
        \caption{Original}
    \end{subfigure}
    \hfill
    \begin{subfigure}[b]{0.24\linewidth}
    \centering
        \includegraphics[width=\textwidth]{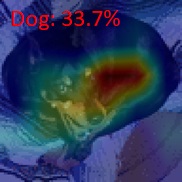}
        \caption{4-bit}
    \end{subfigure}
    \hfill
    \begin{subfigure}[b]{0.24\linewidth}
    \centering
        \includegraphics[width=\textwidth]{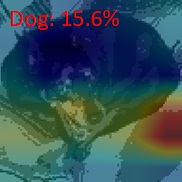}
        \caption{2-bit}
    \end{subfigure}
    \hfill
    \begin{subfigure}[b]{0.24\linewidth}
    \centering
        \includegraphics[width=\textwidth]{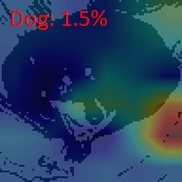}
        \caption{1-bit}
    \end{subfigure}
    
\caption{
Original and MedianCut~\cite{heckbert1982color} color quantized images (top) and corresponding class activation maps~\cite{zhou2016learning} with network confidence for the ground truth class (bottom). In ``Original'' (a), colors are described with 24 bits. When the color space reduces, the focus of the neural network deviates from the informative parts in (a) (correctly recognized), resulting in recognition failures in (b), (c), and (d). 
}

\label{fig:traditional}
\end{figure}

In the literature, a closely related line of work on the interplay between colors and structures is color quantization. Color quantization focuses on generating visually accurate images in restricted color spaces~\cite{heckbert1982color,orchard1991color}. 
This problem is \textit{human-centered}, as it usually prioritizes the visual fidelity or quality for human viewing. 
To this end, traditional color quantization methods minimize the color distortion during quantization. Specifically, they cluster the pixels according to their RGB color values: pixels of similar colors are grouped into the same cluster, and pixels of dis-similar colors are assigned into different clusters. The color quantization process is finalized by assigning an overall color to all pixels for each cluster. 
In this manner, pixels in the color quantized image have RGB color values very similar to the original ones, hence preserving visual fidelity.
See Fig. \ref{fig:traditional} (b) for an example: in a 4-bit color space, the color quantization result is still visually similar to the original image in a 24-bit color space. 

\begin{figure*}
\centering
\includegraphics[width=\linewidth]{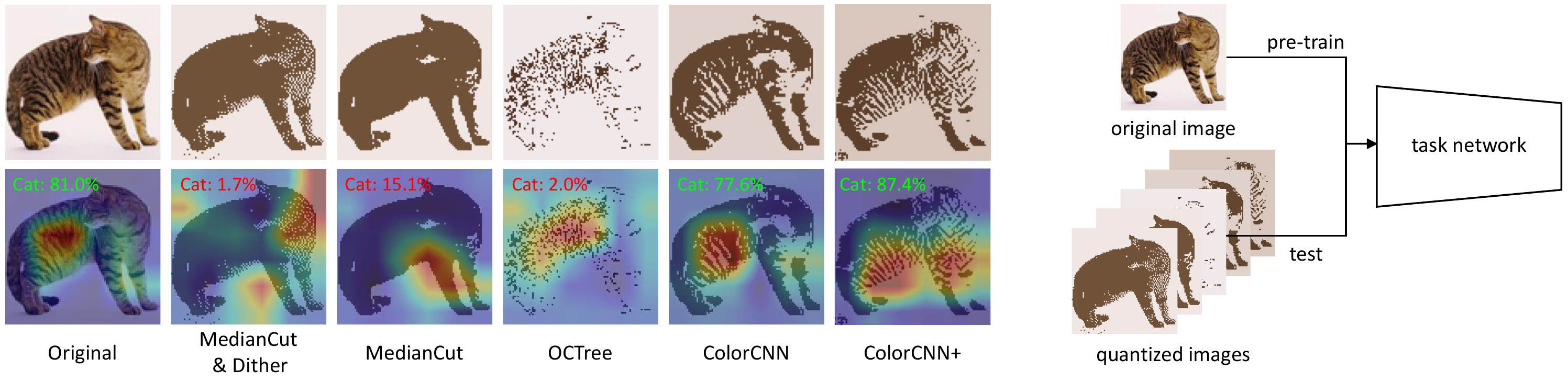}
\caption{
\textbf{Quantization results:} 
``MedianCut \& Dither''~\cite{heckbert1982color,floyd1976adaptive}, ``MedianCut''~\cite{heckbert1982color}, and ``OCTree''~\cite{gervautz1988simple} are traditional color quantization methods; ``ColorCNN'' and ``ColorCNN+'' are the proposed methods. 
\textbf{Evaluation:} 
We evaluate the color quantized images using a classification network pre-trained on the original images. 
\textbf{Comparison:} Traditional methods quantize the original image based on only RGB color values, and hence may lose important shapes and textures, resulting in recognition failures. 
On the other hand, through learning to structure, images quantized by our methods better preserve shapes and textures, and are successfully recognized by the pre-trained classifier. 
}
\label{fig:pre-train_test}
\end{figure*}

Finding key structures for network recognition is orthogonal to traditional color quantization problems, as it is \textit{network-centered}: neural network recognition accuracy is its major focus, instead of human viewing experiences. 
To investigate the key structures for neural network recognition, we further reduce the color spaces to just a few bits, so as to minimize the contribution of colors to the recognition task. In this manner, we can ensure that the neural networks primarily relies on the preserved structures for recognition. 
Natural images usually contain rich colors and structures, and further limiting the color space will inevitably compromise the structure. 
For example, in Fig.~\ref{fig:traditional} (c) and (d), as we reduce the color space sizes all the way down to 1-bit (two colors), most structures vanish. 
When given such images in very limited color spaces, a neural network trained on original natural images might be ineffective, as there are neither accurate colors nor informative structures in such images. 
In this example, with the gradual reduction of the color space, the network fails to find the dog head and then the dog body, leading to recognition failures.

In this work, to identify the key structures, we design a color quantization method, ColorCNN, which learns to preserve informative structure within an image in an end-to-end manner. Through minimizing the classifier loss and improving the recognition accuracy, ColorCNN learns the informative structures for recognition under limited color space. Unlike traditional color quantization methods that rely on RGB color values to preserve accurate colors and visual fidelity for human viewing, ColorCNN exploits semantics to identify and preserve the critical structures for machine perception. 
As shown in Fig.~\ref{fig:pre-train_test}, we evaluate the quantized images with a classifier pre-trained on original images. 
Images quantized by ColorCNN include richer structures like the overall shape, tabby stripes, and the cat head, enabling the classifier to successfully recognize the cat even under extremely limited color spaces.

\begin{figure}
\centering
\includegraphics[width=\linewidth]{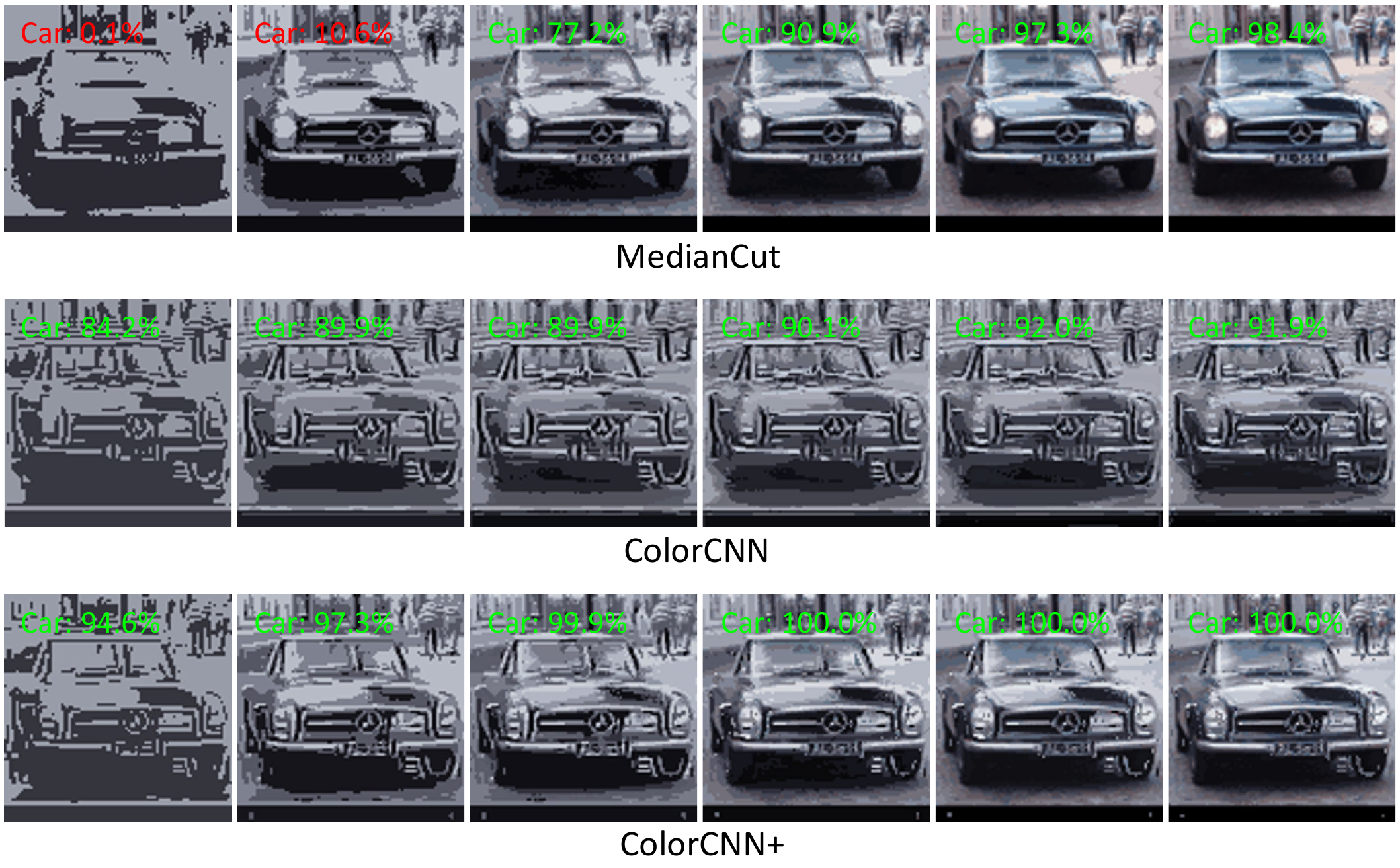}
\caption{Color quantization using traditional clustering-based method (MedianCut \cite{heckbert1982color}), ColorCNN \cite{hou2020learning}, and ColorCNN+. From left to right are results under 1 to 6-bit color spaces. 
In \textbf{small} (1 and 2-bit) color spaces, ColorCNN and ColorCNN+ maintain \textit{key structures}, enabling network recognition. 
In \textbf{large} (5 and 6-bit) color spaces, optimizing directly for classification loss can cause overfitting, and no longer leads to competitive recognition accuracy for ColorCNN. 
ColorCNN+, on the other hand, outputs more visually accurate results with smaller \textit{color distortions} (MedianCut is the most visually accurate). This approach addresses the overfitting issue, and thus gives higher accuracy. 
}
\label{fig:comparison_threeway}
\end{figure}

Color quantization results from ColorCNN (Fig.~\ref{fig:comparison_threeway} mid) show a clear preference towards \textit{key structures} for network recognition over \textit{visual fidelity} with accurate colors, which is quite different from the traditional approaches like MedianCut \cite{heckbert1982color} (Fig.~\ref{fig:comparison_threeway} top). Under small color spaces, \ie, 1-bit or 2-bit, ColorCNN identifies the outline, logo, windshield, and wheels as key structures, and hence, allows for successful recognition. 
However, when the color space size increases, directly optimizing for accuracy can lead to overfitting, and no longer helps the vanilla ColorCNN to achieve competitive accuracy. 
The identified critical structures also provide limited help in such scenarios, as structures can be naturally supported by the large number of colors. 
To address this issue, we design a new architecture in this journal extension, ColorCNN+, that optimizes for not only classification loss, but also visual fidelity under large color spaces (Fig.~\ref{fig:comparison_threeway} bottom). 
As it successfully deals with the training accuracy overfitting issue in vanilla ColorCNN, ColorCNN+ achieves higher recognition accuracy under large color spaces. 

This paper extends our conference version \cite{hou2020learning} in several critical aspects. 
\textbf{First}, the vanilla ColorCNN has to be re-trained for every specific color space size, which brings difficulties to its deployment and real-world usage. In its enhanced version, ColorCNN+, we use a single model to perform color quantization under different color space sizes. 
\textbf{Second}, for larger color spaces, color quantization results from ColorCNN display low visual fidelity with large color distortions, and have dis-satisfactory recognition accuracy. In ColorCNN+, we minimize the color distortion by introducing a novel imitation learning approach, which learns to cluster the colors like traditional color quantization methods. This addresses the previous overfitting issue, and leads to higher visual fidelity (more accurate colors compared to ColorCNN) and competitive recognition accuracy under large color spaces. 
\textbf{Third}, we include more experiments on other classification settings, including multi-label classification and classification on stylized images. The new experiments enable us to compare structure preferences of different networks on different tasks, and provide us deeper insights into how networks recognize an image. 
\textbf{Fourth}, 
we show and discuss how informative structures and accurate colors benefit neural network recognition. 
Specifically, we find that despite their collaboration in large color spaces (which naturally supports more structures), visual fidelity (accurate colors) for \textit{human} viewing and key structures for \textit{network} recognition might contradict each other in small color spaces. 

We demonstrate the effectiveness of ColorCNN and ColorCNN+ quantization on image classification tasks. It is found that both methods outperform traditional ones by a large margin under small color spaces, while ColorCNN+ remains competitive under large color spaces. Such results verify that the preserved patterns from ColorCNNs indeed help network recognition in small color space, and can be considered as key structures for network recognition.
For applications, ColorCNNs stand as competitive image compression methods that enable effective neural network recognition while enjoying very low bitrates. 
 

\section{Related Work}\label{sec:Related_Work}
\textbf{Color quantization.} 
Color quantization~\cite{orchard1991color,deng1999peer,achanta2012slic,deng2001unsupervised,wu1992color} shrinks the color space sizes by grouping similar colors together and represent them with a new color. In this manner, color quantization can reduce image storage while keeping visual similarity to the original image. 
To best preserve visual fidelity, color quantization is usually formulated as a color value clustering problem. Many efficient color clustering methods are introduced, including the popular MedianCut~\cite{heckbert1982color} and OCTree~\cite{gervautz1988simple} algorithms.
Dynamic programming~\cite{wu1992color} and peer group filtering~\cite{deng1999peer} are also investigated for color quantization.
Dithering~\cite{floyd1976adaptive}, which removes visual artifacts by adding a noise pattern, is an optional step for better human viewing experience.  
Color quantization techniques are also applied in segmentation, including JSEG~\cite{deng2001unsupervised}, SLIC~\cite{achanta2012slic}, and more. 

\textbf{Human-centered image compression.} Based on heuristics, many image compression methods are designed for human viewers. 
These methods fall into two categories, lossless compression, \eg, PNG~\cite{boutell1997png}, and lossy compression, \eg, JPEG~\cite{wallace1992jpeg,skodras2001jpeg}. 
Color quantization also falls in the category of lossy compression. Its results, however, can be encoded in a lossless manner. 
The color quantized images can be represented as \textit{indexed color}~\cite{poynton2012digital}, and encoded with Portable Network Graphics (PNG)~\cite{boutell1997png}.

Recently, deep learning methods are introduced to image compression problems. Both recurrent methods~\cite{oord2016pixel,johnston2018improved,toderici2017full} and convolutional methods~\cite{mentzer2019practical,li2018learning,van2016conditional,agustsson2018generative} are investigated.
Some method~\cite{toderici2017full,johnston2018improved} minimize distortion loss under different compression ratio. On the other hand, aiming to achieve a better compression ratio under multiple bitrate settings, some~\cite{balle2016end,balle2018variational} directly optimize rate-distortion instead. 
One possible drawback of these deep methods is that they have a much higher computation cost and need a separate decoder neural network.

\textbf{Network-centered image compression.} 
Traditional or deep-learning based, the aforementioned image compression methods are human-centered. However, in many cases, human-centered methods are not the best choice for neural network tasks. 
Liu \etal~\cite{liu2019machine} points out that for segmentation, human-centered compression is not the best choice for 3D medical images. 
For 2D map data and 3D scene models, network-centered compression methods are designed for localization~\cite{wei2019learning,camposeco2019hybrid}.
Researchers use end-to-end trainable auto-encoder architecture for the map data compression. 
%


\begin{figure*}[t]
\begin{center}
\centering
\includegraphics[width=0.9\linewidth]{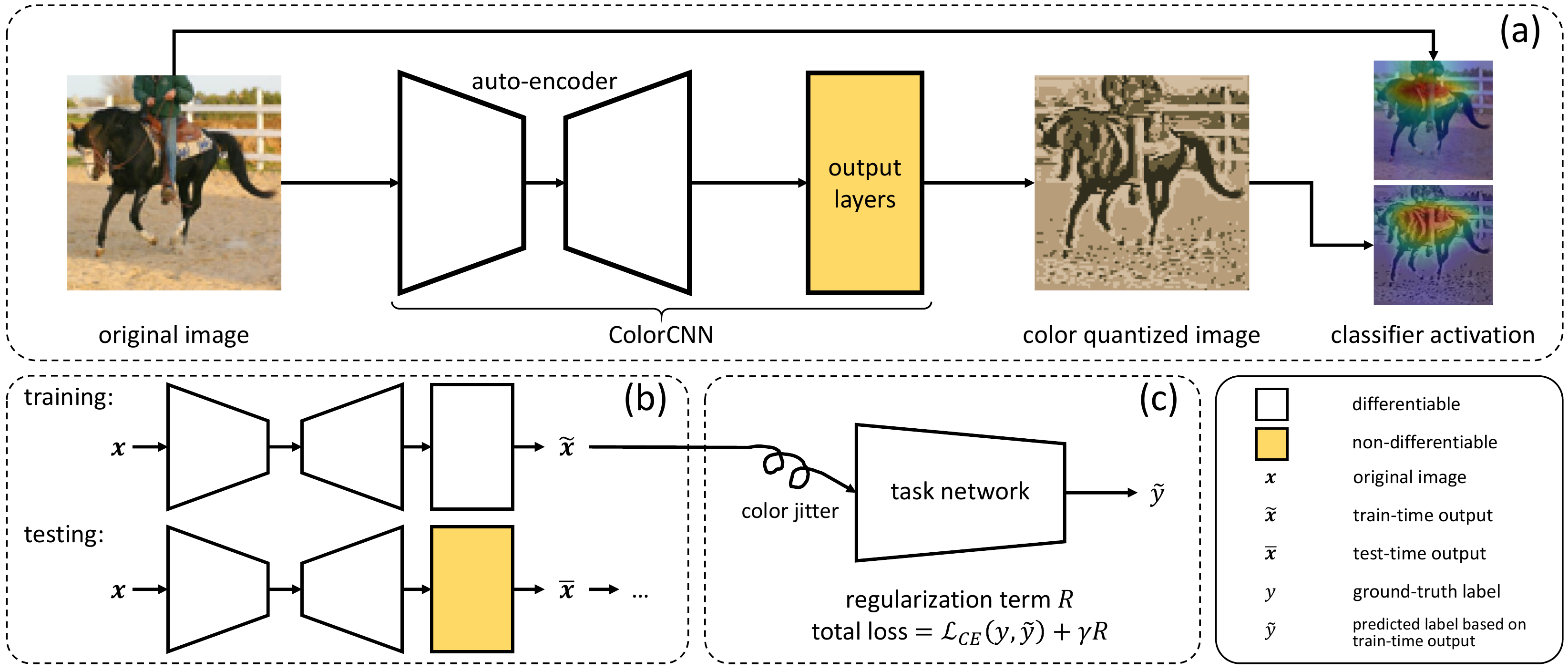}
\end{center}
\caption{Overview of the ColorCNN approach in finding key structures for network recognition. (\textbf{a}): ColorCNN applies network-centered color quantization in limited color space, so as to isolate the contribution of colors to the recognition task. 
(\textbf{b}): We replace the non-differentiable parts with approximations during training. (\textbf{c}): The ColorCNN network is trained with classification loss in an end-to-end manner. Regularization term $R$ and color jitter are introduced to keep the approximation similar to the original network and prevent overfitting. 
}
\label{fig:system}
\end{figure*}

\textbf{Deep clustering.} 
Using neural networks to solve the clustering problem is non-trivial, and stands as a challenging problem itself. 
Self-organizing maps \cite{kohonen1982self} reduce the input dimension by creating a discretized grid as the representation of training samples. More recently, many investigate deep neural networks as a means for dimensionality reduction or representation learning. Many \textit{clustering losses}, including k-means loss \cite{yang2017towards} (distance with k-means cluster center), cluster assignment hardening loss \cite{xie2016unsupervised,li2018discriminatively,aljalbout2018clustering} (promotes more confident cluster assignments), cluster classification loss \cite{hsu2017cnn,wang2016learning} (deems the clusters as classes). For the final \textit{cluster updates}, some adopt the established algorithms like k-means or Agglomerative clustering \cite{hsu2017cnn,yang2016joint,li2018discriminatively,chen2017unsupervised}, while others use neural networks to output cluster centers  \cite{xie2016unsupervised,ghasedi2017deep,yang2017towards,aljalbout2018clustering} or soft cluster assignments  \cite{wang2016learning,hu2017learning}. 
In this work, in order to preserve visual fidelity with color clustering, we investigate an alternative approach in deep clustering. Specifically, we adopt a novel imitation learning loss as the \textit{clustering loss}, and a fully convolutional pipeline for the final \textit{cluster update}.

\section{Methodology}\label{sec:Method}
In this section, we first formulate the learning-to-structure problem mathematically in Section~\ref{secsec:formulation}. Next, to identify and preserve the critical structures in original images, we design ColorCNN architecture and an end-to-end training method (see Fig.~\ref{fig:system}) in Section~\ref{secsec:ColorCNN_structure} and Section~\ref{secsec:colorcnn_training}, respectively. 
Preliminary experiments find that ColorCNN effectively preserve key structures and enable network recognition under small color spaces, but overfits under large color spaces, suffering from poor visual fidelity and undesirable recognition accuracy. 
Identifying such problems in the vanilla ColorCNN design, we then introduce its improved version, ColorCNN+, which can preserve both key structures under small color spaces and visual fidelity (accurate colors) under large color spaces. We introduce the ColorCNN+ network architecture in Section~\ref{secsec:colorcnn+ arch}, and the updated training pipeline in Section~\ref{secsec:colorcnn+_training} and Section~\ref{secsec:colorcnn+_details}. ColorCNN+ also enables us to investigate how key structures and accurate colors contribute to neural recognition, which we take a deeper dive into in the next section. 

\subsection{Problem Formulation}\label{secsec:formulation}




Given an input original image $\bm{x}$ and a color space size $C$, color quantization methods output the color quantized image $\bm{\overline{x}}$, which can be represented and encoded by the color index map $M\!\left(\bm{x}\right)$ and the color palette $T\!\left(\bm{x}\right)$: using the color index map as a lookup table and filling in colors from the color palette, a color quantized image can be reconstructed. 

Our objective then is to construct an image out of a few colors such that the color quantization result can still be correctly recognized by a pre-trained classifier. We consider the following, 
\begin{equation}\label{eq:optim_quantizer}
\mathcal{L}={\mathcal{L}_\text{CE}\left(y, \widetilde{y} \right)+\gamma R},
\end{equation}
as our overall loss function, 
where $\mathcal{L}_\text{CE}\left(\cdot,\cdot\right)$ denotes the cross-entropy classification loss. The variable $y$ denotes the ground truth label for image $x$, and $\widetilde{y}$ denotes the pre-trained classifier output from the color quantized image. $R$ is a regularization term and $\gamma$ denotes its weight.

\subsection{ColorCNN Architecture}\label{secsec:ColorCNN_structure}

\begin{figure}[t]
\begin{center}
\centering
\includegraphics[width=\linewidth]{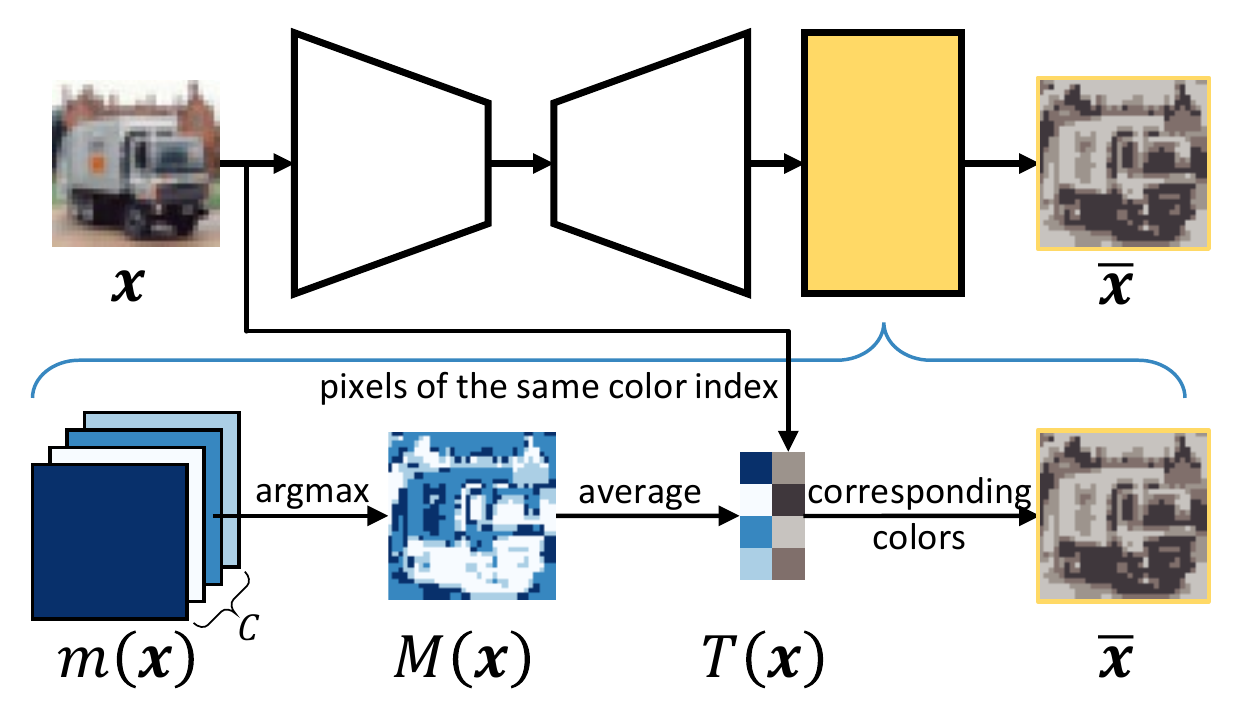}
\caption{ColorCNN architecture (\textbf{test-time}). First, the convolutional layers output a $C$-channel probability map $m\left(\bm{x}\right)$ for $C$ colors. Next, a $1$-channel color index map $M\left(\bm{x}\right)$ is created via the $\argmax$ function. Then, the color palette $T\left(\bm{x}\right)$ is computed as average of all pixels that are of the \textit{same color index}. At last, the color quantized image $\bm{\overline{x}}$ is created via a \textit{table look-up} session. 
}
\label{fig:forward_test}
\end{center}
\end{figure}

We show the ColorCNN architecture in Fig.~\ref{fig:forward_test}. 
Its first component is an U-net~\cite{ronneberger2015u} auto-encoder that identifies the critical and semantic-rich structures.


Secondly, we use depth-wise ($1\times1$ kernel size) convolution layer to create a softmax probability map of each pixel taking one specific color. 
This results in a $C$-channel probability map $m\!\left(\bm{x}\right)$ (softmax over $C$-channel). 

Then, for each input image $\bm{x}$, the $1$-channel color index map $M\!\left(\bm{x}\right)$ is computed as the $\argmax$ over the $C$-channel probability map $m\!\left(\bm{x}\right)$,
\begin{equation}\label{eq:color_index_test}
    M\!\left(\bm{x}\right) = \argmax_{c}{\,m\!\left(\bm{x}\right)}.
\end{equation}

The RGB color palette, $T\left(\bm{x}\right)$, which is of shape $C\times 3$, is computed as average of all pixels that falls into certain quantized color index,
\begin{equation}\label{eq:color_palette_test}
    \left[T\left(\bm{x}\right)\right]_c = \frac{\sum_{\left(u,v\right)}{\left[\bm{x}\right]_{u,v} \bigcdot \mathbb{I}\left(\left[M\left(\bm{x}\right)\right]_{u,v}=c\right)}}{\sum_{\left(u,v\right)}{\mathbb{I}\left(\left[M\left(\bm{x}\right)\right]_{u,v}=c\right)}},
\end{equation}
where $\left[\cdot\right]_i$ denotes the $i$-th element or tensor of the enclosed entity, $\mathbb{I}\left(\cdot\right)$ is the identity function taking value $1$ if its argument is true and $0$ otherwise, and 
$\bigcdot$ denotes component-wise multiplication. 
For pixel $\left(u,v\right)$ in a $W\times H$ image, $\left[x\right]_{u,v}$ denotes the pixel and its RGB value in the input image, and $\left[M\left(\bm{x}\right)\right]_{u,v}$ represents its computed color index. $\left[T\left(\bm{x}\right)\right]_c$ denotes the RGB value for the quantized color $c$. 

Finally, the quantized image $\bm{\overline{x}}$ is created via a table lookup session, which can be represented as,
\begin{equation}\label{eq:reconsturect_test}
    \bm{\overline{x}} = \sum_{c}{\left[T\left(\bm{x}\right)\right]_c\bigcdot \mathbb{I}\left(M\left(\bm{x}\right)=c\right)}.
\end{equation}

By combining Eq.~\ref{eq:color_index_test},~\ref{eq:color_palette_test},~\ref{eq:reconsturect_test}, we finish the ColorCNN forward pass $\bm{\overline{x}} = g\!\left(\bm{x}\right)$ under color space size $C$. 


\subsection{End-to-End Learning} \label{secsec:colorcnn_training}

\subsubsection{Differentiable Approximation}
\label{secsecsec:approximation}
Fig.~\ref{fig:forward_train} shows the differentiable approximation used during training. 
To start with, we remove the $\argmax$ $1$-channel color index map $M\!\left(\bm{x}\right)$ in Eq.~\ref{eq:color_index_test}. Instead, we use the $C$-channel softmax probability map $m\!\left(\bm{x}\right)$.


Next, we change the color palette design that follows. 
For each quantized color, instead of averaging from \textit{pixels of the same color index}, we set its RGB color value $\left[t\left(\bm{x}\right)\right]_c$ as the weighted average over \textit{all pixels}, 
\begin{equation}\label{eq:color_palette_train}
    \left[t\left(\bm{x}\right)\right]_c = \frac{\sum_{\left(u,v\right)}{\left[\bm{x}\right]_{u,v} \bigcdot \left[m\left(\bm{x}\right)\right]_{u,v,c}}}{\sum_{\left(u,v\right)}{\left[m\left(\bm{x}\right)\right]_{u,v,c}}}.
\end{equation}
Here, the $C$-channel probability distribution $\left[m\left(\bm{x}\right)\right]_{u,v}$ for a certain pixel $\left(u,v\right)$ is used as the contribution ratio of that pixel to all $C$ colors. This will result in a slightly different color palette $t\left(\bm{x}\right)$.

Lastly, we change the table look-up process from the original forward pass into a weighted sum. For quantized color with index $c$, we use $\left[m\left(\bm{x}\right)\right]_{c}$ as the intensity of expression over entire image. Mathematically, the train-time quantized image $\bm{\widetilde{x}}$ is computed as,
\begin{equation}\label{eq:reconsturect_train}
    \bm{\widetilde{x}} = \sum_{c}{\left[t\left(\bm{x}\right)\right]_c\bigcdot \left[m\left(\bm{x}\right)\right]_{c}}.
\end{equation}

By combining Eq.~\ref{eq:color_palette_train},~\ref{eq:reconsturect_train}, the forward pass for ColorCNN during training can be formulated as $\bm{\widetilde{x}} = \widetilde{g}\!\left(\bm{x}\right)$. At last, we substitute ${g}\left(\cdot\right)$ with $\widetilde{g}\left(\cdot\right)$ in Eq.~\ref{eq:optim_quantizer} for end-to-end training.


\subsubsection{Overfitting Prevention}\label{secsecsec:overfitting}

\begin{figure}[t]
\begin{center}
\centering
\includegraphics[width=\linewidth]{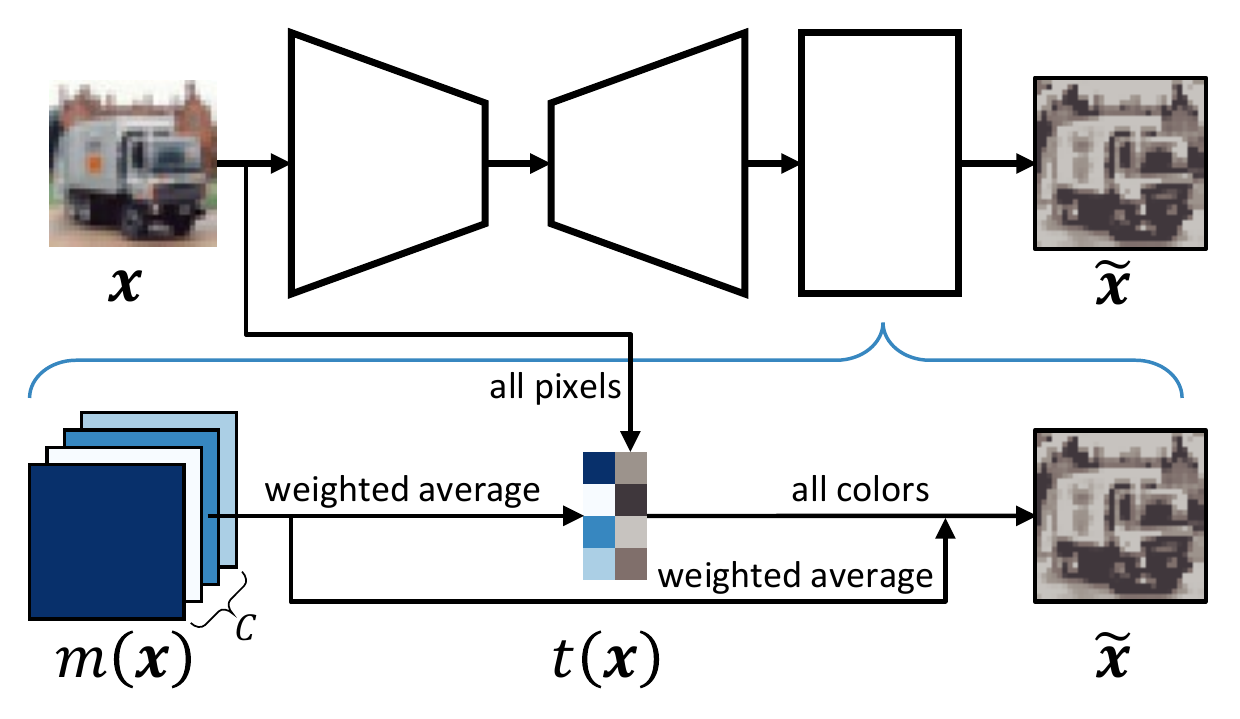}
\end{center}
\caption{The differentiable approximation (\textbf{train-time}). The $C$-channel probability map $m\left(\bm{x}\right)$ is used instead of  the $\argmax$ color index map $M\left(\bm{x}\right)$. Next, the color palette $t\left(\bm{x}\right)$ is adjusted as weighted average over \textit{all} pixels. At last, instead of table look-up, the quantized image $\bm{\widetilde{x}}$ is computed as the weighted average of \textit{all} colors in the color palette. 
}
\label{fig:forward_train}
\end{figure}

\begin{figure}[t!]
    \centering
    \begin{subfigure}[b]{0.45\linewidth}
    \centering
        \includegraphics[width=\textwidth]{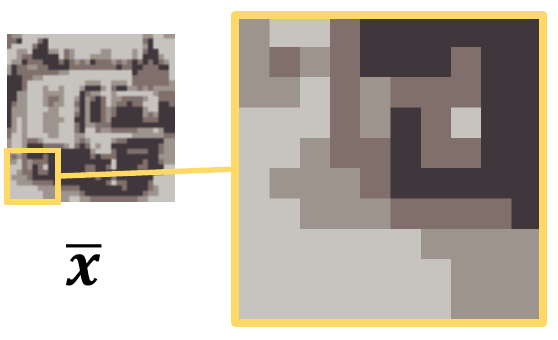}
        \caption{Test-time result}
    \end{subfigure}
    \hfill 
    \begin{subfigure}[b]{0.45\linewidth}
    \centering
        \includegraphics[width=\textwidth]{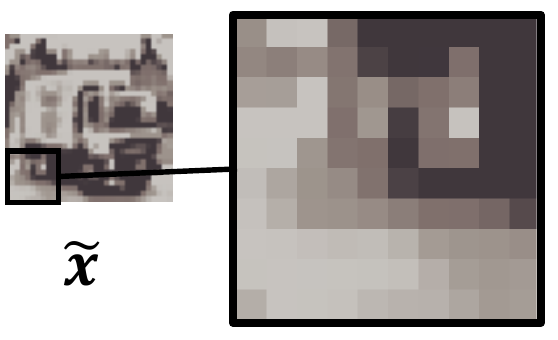}
        \caption{Train-time result}
    \end{subfigure}
\caption{Comparison between test-time result $\bm{\overline{x}}$ and train-time result $\bm{\widetilde{x}}$. Each pixel in $\bm{\widetilde{x}}$ is a weighted average of all colors in its palette. Thus, more colors are introduced. 
}
\label{fig:comparison}
\end{figure}

The two forward pass ${g}\left(\cdot\right)$ and $\widetilde{g}\left(\cdot\right)$ behave very differently. See Fig.~\ref{fig:comparison} for a side-by-side comparison of their outputs.  
The test-time output $\bm{\overline{x}}$ only has $C$ colors, whereas the train-time output $\bm{\widetilde{x}}$ has more than $C$ colors. 
The main reason for this mismatch boils down to the difference between hard (test-time) and soft (train-time) assignments. As shown in Fig.~\ref{fig:forward_test}, the one-hot approach allows influence only from \textit{some} pixels to one quantized color, and from \textit{one} color to any quantized pixel. On the other hand, in Fig.~\ref{fig:forward_train}, with softmax function, \textit{all} pixels influence all colors in the palette, and \textit{all} colors in the palette contribute to each pixel in the output image.

Trained with the approximation $\widetilde{g}\left(\cdot\right)$ instead of the test-time model, the proposed ColorCNN encounters overfitting. 
However, more freedom leads to easy convergence during the training, while the test-time results might still be struggling. 
In the following paragraphs, we introduce regularization terms and data augmentation as means to combat such overfitting.

\begin{figure*}[t]
\centering
\includegraphics[width=0.9\linewidth]{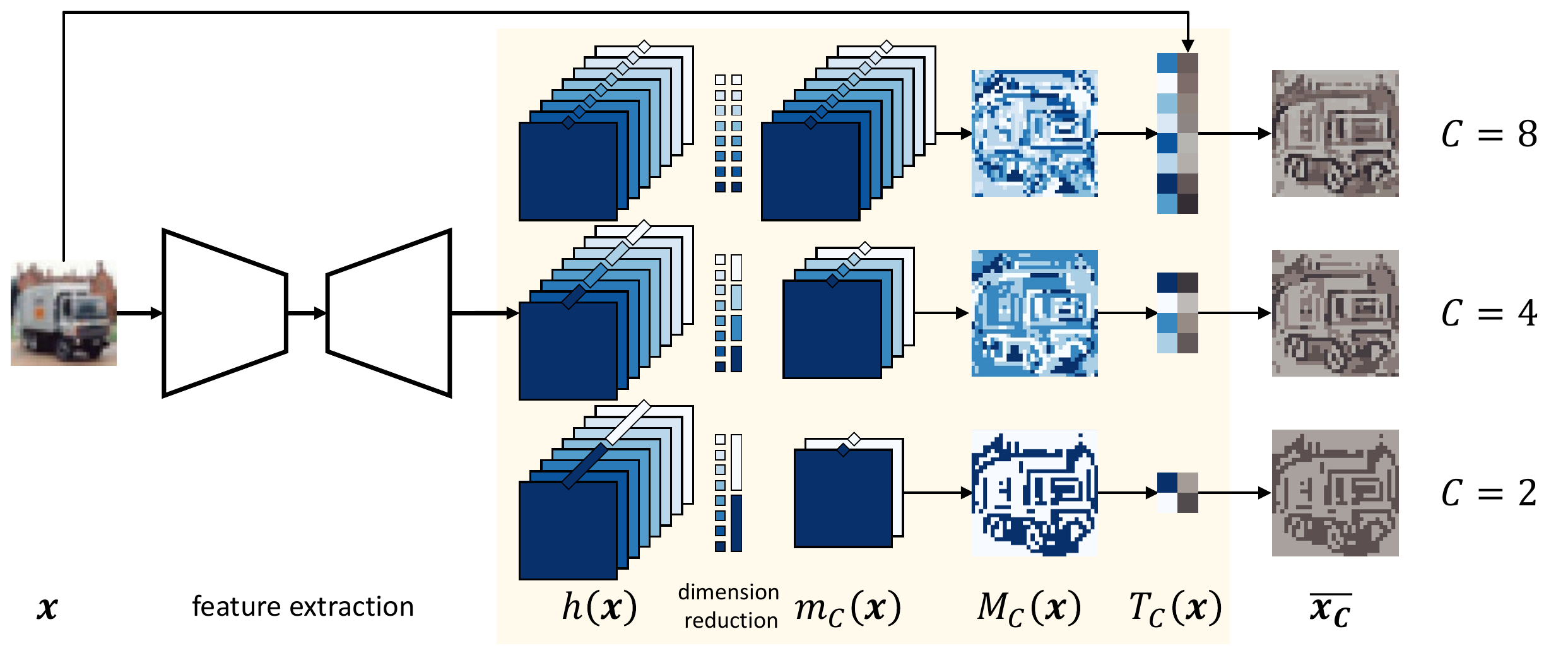}
\caption{ColorCNN+ supports multiple color space sizes in one model. Different from ColorCNN where the color space size $C$ is used to decide the network structure (feature extractor output channel), ColorCNN+ has a fixed structure with auto-encoder output channels of $D$, where $D$ is a predefined number. For multiple color space size support, we reduce the feature map dimension from $D$ to $C$ via channel-wise average pooling to formulate color probability maps $m_C\left(\bm{x}\right)$. 
The rest of the ColorCNN+ including index map $M_C\left(\bm{x}\right)$ and color palette $T_C\left(\bm{x}\right)$ follows the same design as ColorCNN. 
}
\label{fig:colorcnn_plus}
\end{figure*}


\textbf{Regularization.}
ColorCNN only uses image classification loss as supervision. As such, there is no guarantee that each image contains $C$ colors. For this problem, we propose a color appearance regularization term $R_\text{color}$ that encourages all $C$ colors to be selected by at least one pixel in an image. 
For all pixels in an image, we take the maximum probability value along the color channel $c\in\left\{1,...,C\right\}$, and use this value as our regularization term
\begin{equation}\label{eq:regularization_color}
    R = R_\text{color} = -\frac{1}{C}\times \sum_{c}{\max_{\left(u,v\right)}{\left[m\left(\bm{x}\right)\right]_{u,v}}}.
\end{equation}



\textbf{Color jitter.}
In order to prevent overfitting, during training, we add a jitter $\xi\times n$ to the color quantized image $\bm{\widetilde{x}}$ after normalization as a form of data augmentation. The noise $n$ is sampled from a Gaussian distribution $\mathcal{N}\left(0,1\right)$. $\xi$ denotes its weight. This creates a more difficult train-time output for the classifier to recognize, and helps to reduce overfitting. 




The overall approach for ColorCNN is shown in Fig.~\ref{fig:system}.




\subsection{ColorCNN+ Architecture}
\label{secsec:colorcnn+ arch}


ColorCNN helps us analyze the critical structures for neural network recognition and specializes in small color spaces. However, it has two problems. \textbf{First}, ColorCNN architecture only supports one color space size, making it difficult to deploy. \textbf{Second}, ColorCNN overfits in larger color spaces (see Fig.~\ref{fig:comparison_threeway}), leading to less visually accurate outputs (for human viewing) and dis-satisfactory recognition accuracy (for machine perception). In this section, on top of ColorCNN, we build a novel color quantization architecture, ColorCNN+, that supports multiple color space size settings in one model, while also capable of outputting results with high visual fidelity and minimal color distortions via deep clustering. 
Such improvements allows for not only easier real-world application, but also more discussion and deeper insight on informative structures and visual fidelity (accurate colors).  

ColorCNN+ builds on the architecture of ColorCNN. An illustration of the ColorCNN+ architecture is shown in Fig.~\ref{fig:colorcnn_plus}. Like ColorCNN, it still adopts a U-net auto-encoder as its feature extractor. 
After that, rather than directly outputting a $C$-channel color probability map as in vanilla ColorCNN, in its enhanced version, we \textbf{first} introduce a lower dimension \textit{bottleneck layer}, so as to limit the excessive information and better combat the overfitting issue. \textbf{Next}, this bottleneck layer is followed by a $D$-channel feature map $h\left(\bm{x}\right)$, where $D$ is a relatively large (larger than color space sizes in our experiment) fixed number. This $D$-channel feature map $h\left(\bm{x}\right)$ serves as building blocks of the color probability map, and we can \textbf{then} create different color probability maps $m_C\left(\bm{x}\right)$ for different color space sizes $C$ by reducing the dimension form the fixed number $D$ to the color space size $C$. This dimension reduction is achieved via channel-wise average pooling, 
\begin{align}\label{eq:color_index_from_dimension}
    \left[h_C\left(\bm{x}\right)\right]_{c} &=  \floor*{\frac{D}{C}}\sum_{d=\floor*{\frac{c-1}{C}\times D}}^{\floor*{\frac{c}{C}\times D}}{\left[h\left(\bm{x}\right)\right]_{d}},
\end{align}
where $h_C\left(\bm{x}\right)$ is the dimension reduction result of channel-wise average pooling over the $D$-channel feature map $h\left(\bm{x}\right)$. 

\textbf{Following that}, we adopt a new design when creating the $C$-dimensional color probability map $m_C\left(\bm{x}\right)$. In vanilla ColorCNN, during training, all pixels in the original image $\bm{x}$ contribute to the color palette $t_C\left(\bm{x}\right)$ and quantized image $\bm{\widetilde{x}}_C$, as long as the color probability map $m\left(\bm{x}\right)$ is non-zero.
Compared to the test-time pipeline where only the pixels belonging to a certain quantized color contributes to the color palette and the final output, the training pipeline has significantly higher degree of freedom and is prone to overfitting.
As such, in ColorCNN+, we modify our design of color probability map $m_C\left(\bm{x}\right)$ by allowing up to \textit{top-$K$ nonzero terms}: at each pixel, $m_C\left(\bm{x}\right)$ is assigned with the softmax probability $\sigma\left(\cdot\right)$ over $K$ elements for channels that have top-$K$ elements of $h_C\left(\bm{x}\right)$, or $0$ for other channels,
\begin{align} \label{eq:color_prob_map}
    \left[m_C\left(\bm{x}\right)\right]_{c} &= \begin{cases}
               \sigma\left(\left[h_C\left(\bm{x}\right)\right]_{c}\right), \;\; &\text{if}\;\; c\in\topk_K{\left[h_C\left(\bm{x}\right)\right]_{c}}, \\
                0, \;\;\;\; &\text{else}.
            \end{cases}
\end{align}

\textbf{The rest of} ColorCNN+ forward pass follows the original design in ColorCNN.
To start with, we create the $1$-channel color index map $M_C\left(\bm{x}\right)$ with Eq.~\ref{eq:color_index_test}. The color palette $T_C\left(\bm{x}\right)$ is then created via Eq.~\ref{eq:color_palette_test}. At last, the $C$-color quantized image $\overline{\bm{x}}_C$ is generated via Eq.~\ref{eq:reconsturect_test}.
Combining all components, we have the forward pass of ColorCNN+ $\overline{\bm{x}}_C=g\!\left(\bm{x}, C\right)$.


\subsection{Deep Clustering in ColorCNN+}\label{secsec:colorcnn+_training}

In order to address the previous issue in ColorCNN where the larger color spaces are not effectively utilized (directly optimizing for train-time accuracy can lead to overfitting, see Fig.~\ref{fig:comparison_threeway}), ColorCNN+ learns from clustering-based traditional color quantization methods to maintain visual fidelity. 
In fact, we believe that promoting the higher visual quality also allows for easier network recognition under larger color space, as the large color spaces naturally allows for more structures, making it no longer the bottleneck for recognition accuracy. 
This deep clustering support also allows for more discussion how informative structures and accurate colors benefit neural network recognition.




\begin{figure}[t]
\begin{center}
\centering
    \begin{subfigure}[b]{\linewidth}
    \centering
        \includegraphics[width=\textwidth]{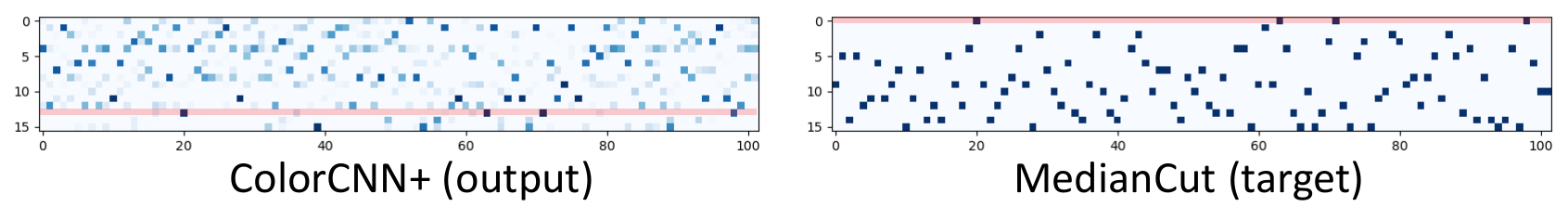}
        \caption{Cluster classification loss (previous work \cite{hsu2017cnn,wang2016learning})}
    \end{subfigure}
    \begin{subfigure}[b]{\linewidth}
    \centering
        \includegraphics[width=\textwidth]{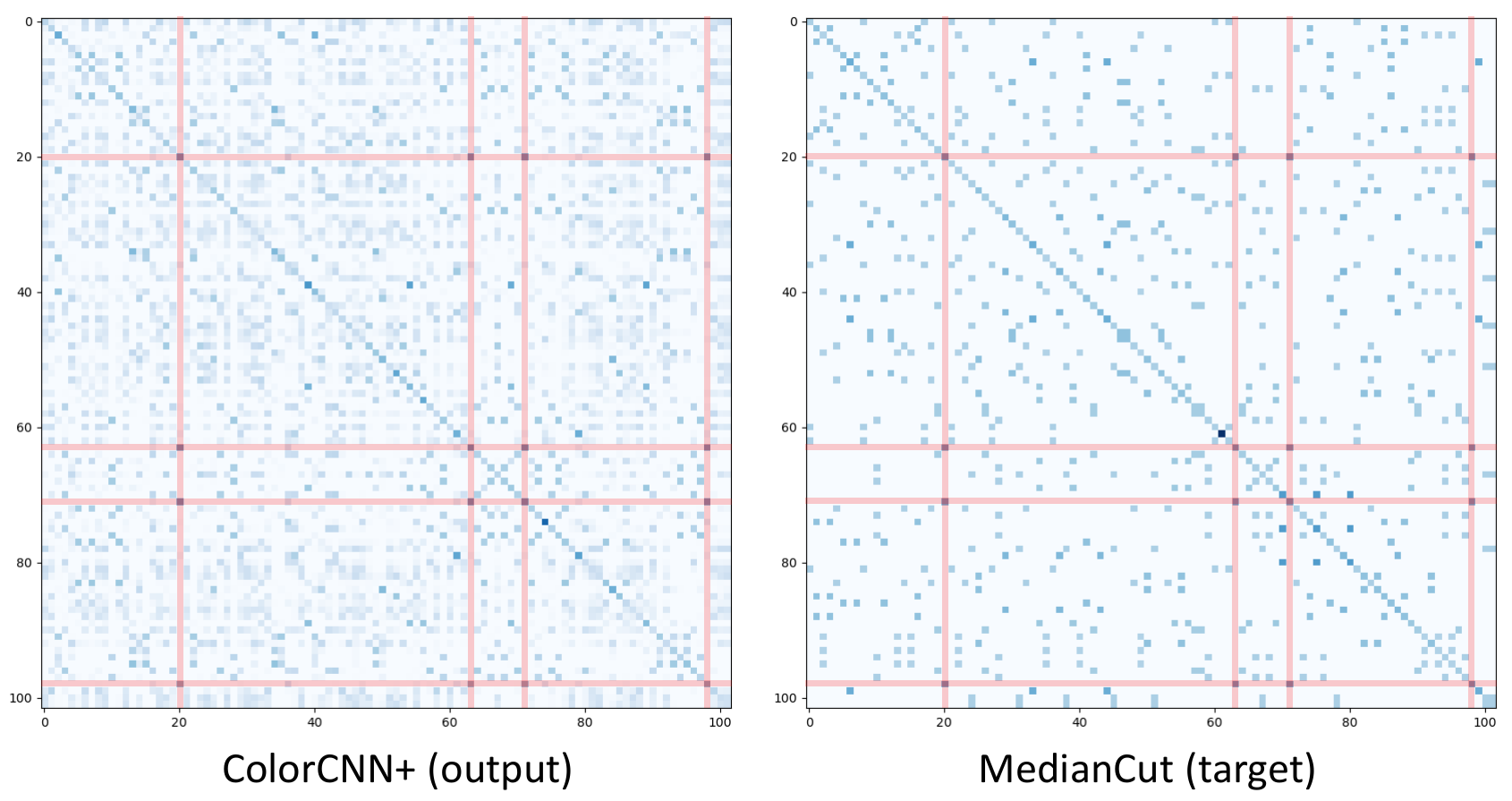}
        \caption{Pixel-wise relationship preserving loss (proposed)}
    \end{subfigure}
\caption{Clustering loss computation for $C=16$ quantized colors over $N=102$ sampled pixels. Highlighted are the same group of pixels with different cluster indices given by different cluster updates. 
}
\label{fig:cluster_loss}
\end{center}
\end{figure}

\subsubsection{Fully Convolutional Cluster Update}

ColorCNN+ aims to preserve both key structures in small color spaces and accurate colors in large color spaces.
It is desired that the network can output an \textit{overall} decision rather than having separate branches for decisions under different scenarios. 

As for the clustering problem formulation, ColorCNN+ neither introduces a separate clustering step as  \cite{yang2016joint,hsu2017cnn,li2018discriminatively,huang2014deep,chen2017unsupervised}, nor outputs the cluster centers \cite{xie2016unsupervised,ghasedi2017deep,yang2017towards,aljalbout2018clustering} and assign cluster indices according to feature distance. Instead, it follow the same per-pixel classification formulation as in vanilla ColorCNN: estimating a soft assignment for each pixel to one of $C$ clusters. 
This creates a fully convolutional pipeline that does not need additional steps for clustering updates. 
Furthermore, we can easily incorporate the clustering results (for visual fidelity and accurate colors in large color spaces) into the current semantic-based results (for key structures in small color spaces) and make an \textit{overall} decision with the color probability map $m_C\left(\bm{x}\right)$.

\subsubsection{Imitation Learning}

For the ColorCNN+ architecture to learn color clustering, we introduce a new imitation learning process. Since we directly estimate the soft cluster assignment, the k-means loss  \cite{yang2017towards} (requires a separate clustering step) and cluster assignment hardening loss \cite{xie2016unsupervised,li2018discriminatively,aljalbout2018clustering} (when estimating cluster centers) from previous works are not suitable. This leaves us with the cluster classification loss \cite{hsu2017cnn,wang2016learning} (for soft cluster assignment estimation)  as the only candidate in the literature, which treat the cluster index as class and formulate deep clustering as a cluster classification problem. 
However, this approach also has its own problem. Unlike classification problems where the classes are fixed, in clustering problems, the cluster indices are randomly generated and cluster indices from different cluster updates (ColorCNN+ as output or MedianCut as target) do not match: as shown in Fig.~\ref{fig:cluster_loss} (a), the two highlighted clusters, despite having different cluster indices, consist of the same pixels. This can cause large cluster classification loss, whereas in fact minimal loss should come from these pixels as the two decisions are similar.

In this case, we investigate alternative choices that do not rely on the strict correspondence between cluster indices. Inspired by works in related fields that preserve batch-wise, pixel-wise, or channel-wise relationships on the \textit{feature maps} \cite{tung2019similarity,li2020semantic,hou2021visualizing}, in this work, we compute the pixel-wise self correlations between \textit{soft cluster assignments} $m_{C}\left(\bm{x}\right)$ for clustering supervision. 
As shown in Fig.~\ref{fig:cluster_loss} (b), the pixel-wise relationship preserving loss only measures relationships between cluster assignments at each pixel. Unlike the cluster index, the pixel arrangements are always the same between different cluster updates, making it a more suitable supervision for deep clustering.

Given ColorCNN+ soft assignment $m_{C}\left(\bm{x}\right)$ and MedianCut one-hot assignment $m^*_{C}\left(\bm{x}\right)$, we \textbf{first} sample $N$ pixels  over $H\times W$ pixels to produce $\mathcal{M},\mathcal{M}^*\in  \mathbb{R}^{C \times N}$. \textbf{Next}, we compute the pixel-wise self correlations,
\begin{align}
\label{eq:gram}
     A = \mathcal{M}\cdot\mathcal{M}^T, \;
     A^* = \mathcal{M}^*\cdot \left(\mathcal{M}^*\right)^T,
\end{align}
where $A , A^* \in \mathbb{R}^{N \times N}$. We \textbf{then} normalize them row-wise,
\begin{align}
\label{eq:l2norm}
     \widetilde{A}_{\left[i,:\right]} = \frac{A_{\left[i,:\right]}}{\left\|A_{\left[i,:\right]} \right\|_2}, \;
     \widetilde{A}^*_{\left[i,:\right]} = \frac{A^*_{\left[i,:\right]}}{\left\|A^*_{\left[i,:\right]} \right\|_2},
\end{align}
where $\left[i,:\right]$ indicates the $i$-th row in a matrix. 
\textbf{At last}, we define the relationship preserving loss as mean square error (MSE) between the normalized self correlation matrices,
\begin{align}
\label{eq:relationship}
     \mathcal{L}_\text{RP} = \frac{1}{N}  \left\|\widetilde{A} - \widetilde{A}^* \right\|_F^2,
\end{align}
where $\left\|\cdot\right\|_F$ denotes the Frobenius norm (entry-wise $\mathcal{L}_2$ norm for matrix). 

Overall, we re-write the loss function in Eq.~\ref{eq:optim_quantizer} as
\begin{align}
\label{eq:combine_loss}
     \mathcal{L} = \mathcal{L}_\text{CE}+\lambda \mathcal{L}_\text{RP} + \gamma R,
\end{align}
where $\lambda$ is a hyper-parameter for the clustering loss weight.

\subsection{Training Details in ColorCNN+}
\label{secsec:colorcnn+_details}


\subsubsection{Regularization}

In addition to the color appearance regularization term $R_\text{color}$ for ColorCNN (see Eq.~\ref{eq:regularization_color}), for ColorCNN+, we introduce two new regularization terms. A information entropy regularization term $R_\text{info}$ to encourage more evenly-distributed color choices; and a confidence regularization term $R_\text{conf}$ to balance out $R_\text{info}$ and increase confidence for the color probability map $m_C\left(\bm{x}\right)$.


\textbf{Information entropy as regularization.}
In order to maximize the information carried by the image so that the classifier can successfully recognise it, we aim to maximize the \textit{information entropy}~\cite{shannon1948mathematical} of the color quantized image. For the color quantization task, we want the number of pixels that take each of the $c\in\left\{1,...,C\right\}$ quantized colors to be evenly-distributed,
\begin{equation}\label{eq:regularization_info}
    R_\text{info} = -\mathcal{H}\left(\frac{1}{H\times W}\sum_{u,v}{\left[m_C\left(\bm{x}\right)\right]_{u,v}}\right),
\end{equation}
where $\mathcal{H}\left(\cdot\right)$ denotes the entropy function over the color channel $c\in\left\{1,...,C\right\}$. 


\textbf{Confidence as regularization.}
One side-effect of our information entropy regularization $R_\text{info}$ is that it also makes the color probability map $m_C\left(\bm{x}\right)$ less like one-hot. 
In light of this, we propose a confidence regularization term, which promotes the color probability map to be more confident at every pixel. We also use the information entropy function $\mathcal{H}\left(\cdot\right)$ to calculate this confidence regularization term,
\begin{equation}\label{eq:regularization_conf}
    R_\text{conf} = \frac{1}{H\times W}\sum_{u,v}{\mathcal{H}\left(\left[m_C\left(\bm{x}\right)\right]_{u,v}\right)}.
\end{equation}
This confidence regularization $R_\text{conf}$ focuses on \textit{pixel-wise} color distribution, whereas the information regularization $R_\text{info}$ focuses on \textit{image-wise} color distribution. 

Combining all three components, we have the final regularization term for ColorCNN+,
\begin{equation}\label{eq:regularization_comb}
    R = R_\text{color}+\alpha R_\text{info} + \beta R_\text{conf},
\end{equation}
where $\alpha$ and $\beta$ specify the ratio for the three components. 

\subsubsection{Data Augmentation}

To better train ColorCNN+, we make some adjustments to the data augmentations. Most importantly, we move random cropping from before ColorCNN+ to after ColorCNN+. If applied before the quantization network, ColorCNN+ has to waste some of its network capacity on dealing with the black bars artifacts. When moved to after quantization network, such augmentation can still add diversity to the classifier training data and help to combat overfitting. 
In addition, we also introduce some more augmentation after the quantization network, in addition to the color jitter and random cropping, including random erasing \cite{zhong2020random}, rotation, and horizontal flipping.


\subsubsection{Task Selection}
\label{secsecsec:task_selection}

For each forward pass in training, we must specify a color space size $C\in\left\{1,...,64\right\}$ (as we consider at most 64 colors in our experiments) for ColorCNN+. We refer to this selecting different color space size for ColorCNN+ training as task selection. Throughout the entire training process, we select different tasks (color space sizes) to give ColorCNN+  sufficient supervision for different color space sizes it could encounter during testing. 

The pace of changing the task is a very important aspect of successful training of ColorCNN+. Similarly, Wu \etal \cite{wu2020multigrid} study the task selection problem for video understanding, where different tasks refer to different video resolutions and batch sizes. It is found that neither too fast nor too slow the pace in task changes lead to high performance. In our color space size selection task, we also witness a similar phenomenon, where changing task per-batch or per-epoch both lead to inferior performance. As such, we empirically choose an intermediate pace in task change for ColorCNN+ training at once every 20 batches.




\begin{table}[t!]
\caption{Classification accuracy (\%)  of different networks on different datasets.}
\label{tab:classifier}
\centering
\resizebox{\linewidth}{!}{
\begin{tabular}{l|c|c|c|c}
\toprule
         & \multicolumn{1}{c|}{CIFAR10} & \multicolumn{1}{c|}{CIFAR100} & \multicolumn{1}{c|}{STL10} & \multicolumn{1}{c}{Tiny200} \\ \hline
AlexNet~\cite{krizhevsky2012imagenet}  & 86.9                         & 62.1                          & 73.6                       & 50.9                                   \\ \hline
VGG16~\cite{simonyan2014very}    & 93.5                         & 73.1                          & 81.5                       & 62.8                                   \\ \hline
ResNet18~\cite{he2016deep} & 94.6                         & 76.4                         & 83.4                       & 65.5                                   \\ \bottomrule
\end{tabular}
}
\end{table}

\begin{figure}[t]
    \centering
    \begin{subfigure}[b]{\linewidth}
    \centering
        \includegraphics[width=\textwidth]{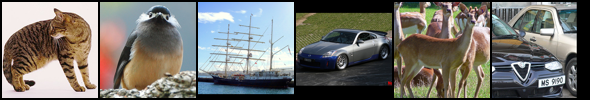}
        \caption{Original}
    \end{subfigure}
    \begin{subfigure}[b]{\linewidth}
    \centering
        \includegraphics[width=\textwidth]{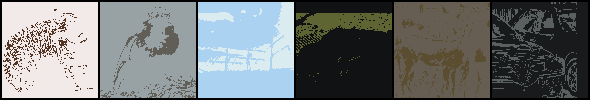}
        \caption{OCTree}
    \end{subfigure}
    \begin{subfigure}[b]{\linewidth}
    \centering
        \includegraphics[width=\textwidth]{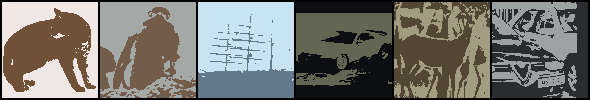}
        \caption{MedianCut}
    \end{subfigure}
    \begin{subfigure}[b]{\linewidth}
    \centering
        \includegraphics[width=\textwidth]{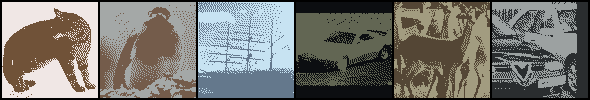}
        \caption{MedianCut \& Dithering}
    \end{subfigure}
    \begin{subfigure}[b]{\linewidth}
    \centering
        \includegraphics[width=\textwidth]{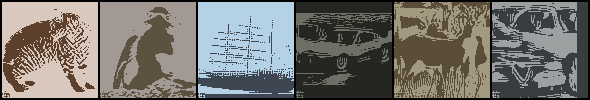}
        \caption{ColorCNN+ with AlexNet}
    \end{subfigure}
    \begin{subfigure}[b]{\linewidth}
    \centering
        \includegraphics[width=\textwidth]{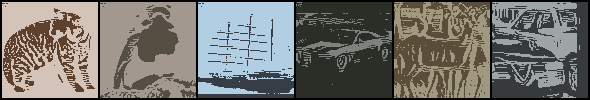}
        \caption{ColorCNN+ with VGG16}
    \end{subfigure}
    \begin{subfigure}[b]{\linewidth}
    \centering
        \includegraphics[width=\textwidth]{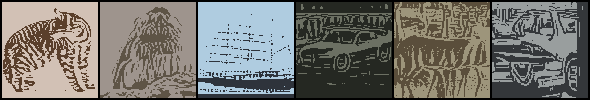}
        \caption{ColorCNN+ with ResNet18}
    \end{subfigure}
\caption{1-bit color quantization results. ColorCNN+ identifies both \textit{textures} and \textit{shapes} as critical structures for neural network recognition. Also, different classifiers prefer different patterns. For more discussion, 
please see Section~\ref{secsec:discussion}.
}
\label{fig:1-bit-batch}
\end{figure}

\begin{figure*}
\centering
    \begin{subfigure}[b]{0.095\linewidth}
    \centering
        \includegraphics[width=\textwidth]{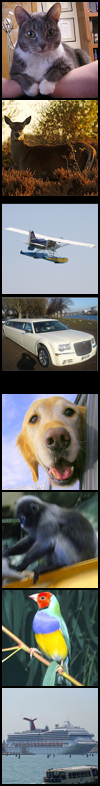}
        \caption{}
    \end{subfigure}
    \begin{subfigure}[b]{0.095\linewidth}
    \centering
        \includegraphics[width=\textwidth]{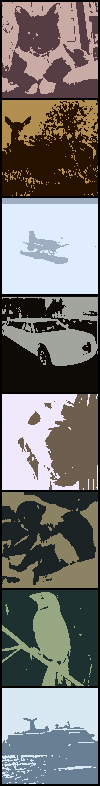}
        \caption{}
    \end{subfigure}
    \begin{subfigure}[b]{0.095\linewidth}
    \centering
        \includegraphics[width=\textwidth]{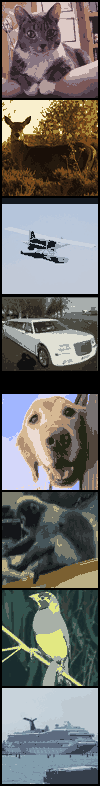}
        \caption{}
    \end{subfigure}
    \begin{subfigure}[b]{0.095\linewidth}
    \centering
        \includegraphics[width=\textwidth]{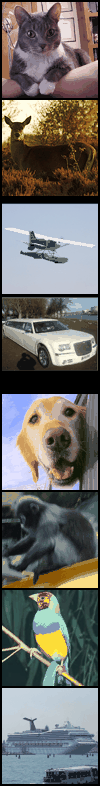}
        \caption{}
    \end{subfigure}
    \begin{subfigure}[b]{0.095\linewidth}
    \centering
        \includegraphics[width=\textwidth]{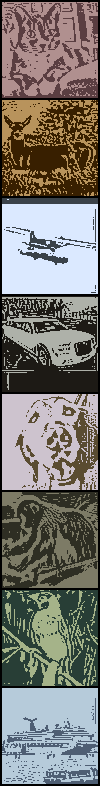}
        \caption{}
    \end{subfigure}
    \begin{subfigure}[b]{0.095\linewidth}
    \centering
        \includegraphics[width=\textwidth]{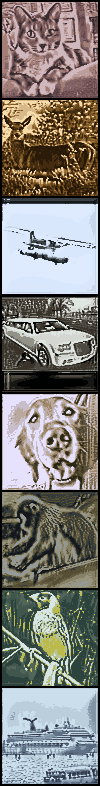}
        \caption{}
    \end{subfigure}
    \begin{subfigure}[b]{0.095\linewidth}
    \centering
        \includegraphics[width=\textwidth]{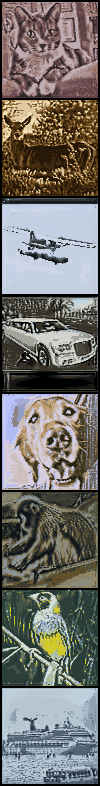}
        \caption{}
    \end{subfigure}
    \begin{subfigure}[b]{0.095\linewidth}
    \centering
        \includegraphics[width=\textwidth]{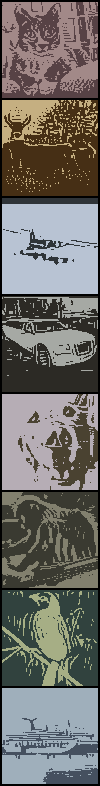}
        \caption{}
    \end{subfigure}
    \begin{subfigure}[b]{0.095\linewidth}
    \centering
        \includegraphics[width=\textwidth]{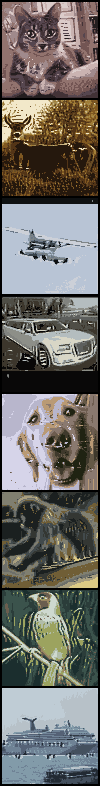}
        \caption{}
    \end{subfigure}
    \begin{subfigure}[b]{0.095\linewidth}
    \centering
        \includegraphics[width=\textwidth]{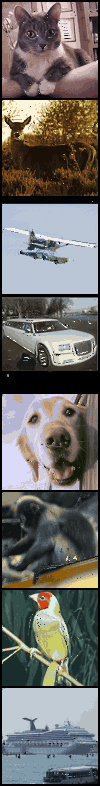}
        \caption{}
    \end{subfigure}
    
\caption{Color quantization results under different color space sizes. Column (a) shows the original images. Column (b)-(d), (e)-(g), and (h)-(j) show MedianCut \cite{heckbert1982color}, ColorCNN, and ColorCNN+ results for 1-bit, 3-bit, and 5-bit color spaces, respectively. 
ColorCNN prioritizes the informative structures over visual fidelity (accurate colors) on both small and large color spaces. 
In comparison, ColorCNN+ identifies and preserves the informative structures under small color spaces, and outputs visually accurate results under large color spaces. For more quantitative results, see Fig.~\ref{fig:accuracy_curves}.
}
\label{fig:colorcnn+_visualization}
\end{figure*}

\section{Experiments}\label{sec:experiment}
\subsection{Experimental Setup}\label{sec:Experiment_Setup}

\textbf{Datasets.} We evaluate on four single-label classification datasets,
one multi-label classification dataset, and one stylized image dataset. Single-label classification is the default task if not specified. 

For single-label classification, we use the following four datasets. 
CIFAR10 and CIFAR100 datasets~\cite{krizhevsky2009learning} include 10 and 100 classes of general objects, respectively. STL10 dataset~\cite{coates2011analysis} contains 10 image classes but with a higher resolution. Tiny-imagenet-200 dataset~\cite{le2015tiny} is a subset of ImageNet-1K dataset~\cite{deng2009imagenet}. It has 200 classes of medium resolution images. 
We omitted experiments on the full ImageNet-1K dataset due to its high complexity. 

For multi-label classification with multiple salient objects, we introduce Pascal VOC 2012 dataset \cite{everingham2010pascal}. 

Stylized ImageNet includes images in arbitrary artistic styles \cite{geirhos2018imagenet}. We create a stylized dataset, Style14, that consists of 14 classes from the full ImageNet-1k dataset \cite{deng2009imagenet}.



\textbf{Classification networks.} We choose AlexNet~\cite{krizhevsky2012imagenet}, VGG16~\cite{simonyan2014very} and ResNet18~\cite{he2016deep} for classification network. 
All classifier networks are trained for $60$ epochs with a batch size of $128$, except for STL10, where we set the batch size to $32$. We use an SGD optimizer with a momentum of $0.9$, L2-normalization of $5\times 10^{-4}$. We choose the 1cycle learning rate scheduler~\cite{smith2019super} with a peak learning rate of $0.1$. We report the classification accuracy on for single-label classification datasets in Table~\ref{tab:classifier}. 

Multi-label classification uses binary cross entropy loss for training, and stylized image classification is trained in the same manner as single-label classification on natural images. We use ResNet18 architecture for both settings. Their performance can be found in Table~\ref{tab:voc} and Table~\ref{tab:style}.

\begin{figure*}[t]
\begin{center}
\centering
\includegraphics[width=\linewidth]{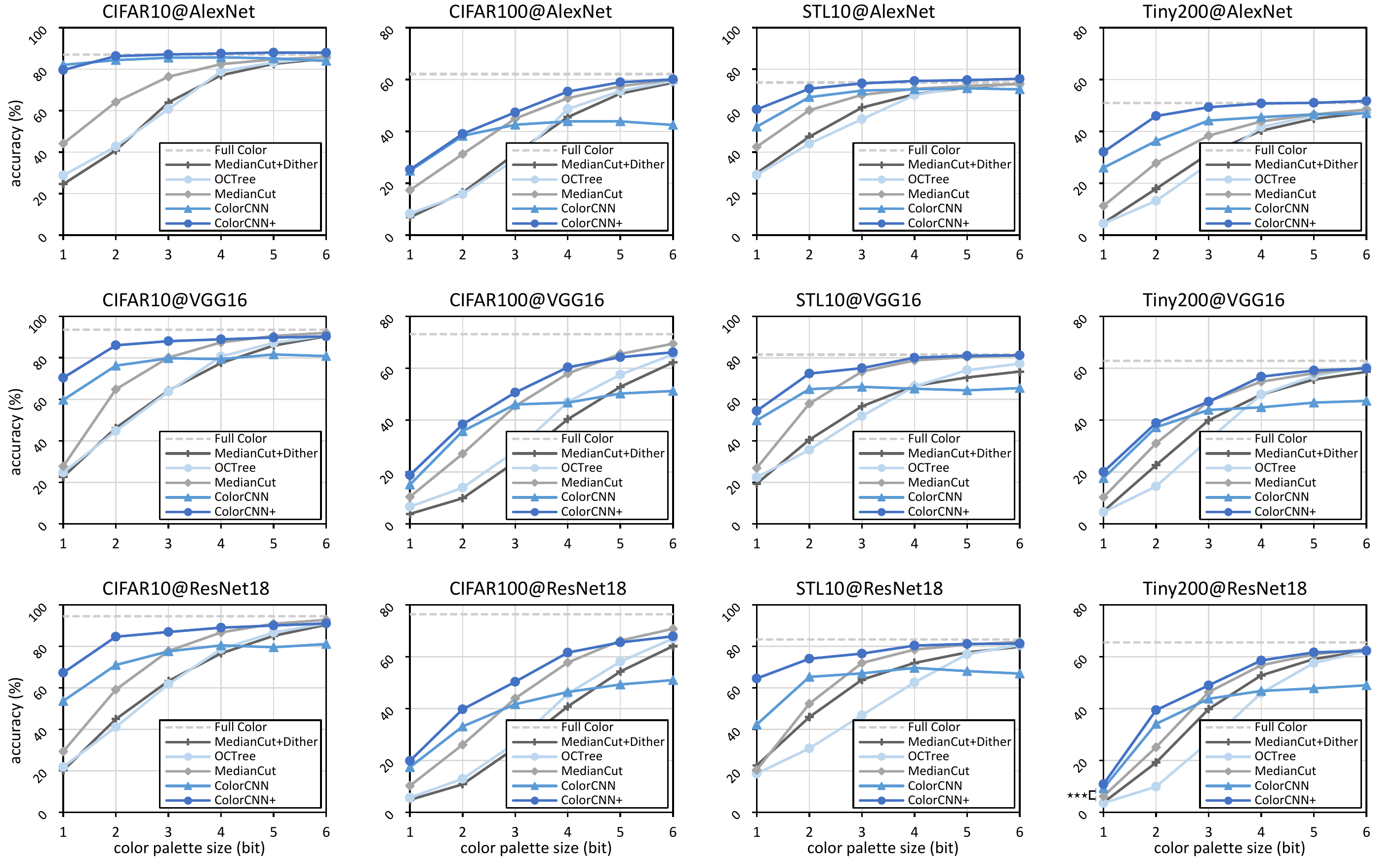}
\caption{Classification accuracy of color quantized images on four datasets with three networks. Compared to traditional methods, both ColorCNNs are significantly better under small color spaces, while ColorCNN+ remains competitive under large color spaces.
$\star\star\star$ means that the accuracy difference is \textbf{statistically very significant} (\ie, $p\text{-value} <0.001$).
}
\label{fig:accuracy_curves}
\end{center}
\end{figure*}

\textbf{ColorCNN.} We train ColorCNN on top of the original image pre-trained classifier. We set the hyper-parameters as follows. For the regularization and color jitter weight, we set $\gamma=1$ and $\xi=1$. 
We also normalize the quantized image by $4\times$ the default variance of the original images, so as to enable easier training. 
The gradient descent optimizer runs 60 epochs with a batch size of $128$. Similar to classification networks, we also reduce the batch size to $32$ on the STL10 dataset. The SGD optimizer for ColorCNN training is the same as for classifier networks. For the learning rate scheduler, we choose Cosine-Warm-Restart~\cite{loshchilov2016sgdr} with a peak learning rate at $0.01$, minimal learning rate at $0$, and uniform restart period of $20$ epochs.

\textbf{ColorCNN+.} We make the following modifications to ColorCNN+ over ColorCNN. Architecture-wise, we add a $16$-dimensional bottleneck layer before the $D=256$-dimensional feature map, and set the number of non-zero terms in color probability map as $K=4$. For the hyper-parameters, we set the weight for the clustering loss $\lambda=3$, and the weights for regularization terms as $\alpha=1$ and $\beta=1$. As for the number of pixels considered in our clustering loss, we set $N$ as $0.3\times$ the number of pixels for memory concerns. 
We train ColorCNN+ for 300 epochs and keep other hyper-parameters the same.

All experiments are run on one RTX-3090 GPU. 

\subsection{Evaluation of ColorCNNs}\label{sec:Evaluation_of_ColorCNN}

We evaluate ColorCNNs and compare them against MedianCut~\cite{heckbert1982color}, OCTree~\cite{gervautz1988simple}, and MedianCut with dithering~\cite{floyd1976adaptive}. 

\subsubsection{Visualization}
\label{secsecsec:visual}

\textbf{Visualization results in \textit{small} color spaces.} As the color space size shrinks, pixels struggle to support structures. Under such scenarios, keeping critical and informative structures becomes the key to successful network recognition. In a minimum 1-bit color space (Fig.~\ref{fig:1-bit-batch} and Fig.~\ref{fig:colorcnn+_visualization}), under the same setting, traditional color quantization method like MedianCut merely keeps some \textit{outlines} of the objects and usually loses fine-grained \textit{textures}. 
In comparison, it is identified by ColorCNNs that both both \textit{shapes} (\eg, boat hull and vehicle outline in Fig.~\ref{fig:1-bit-batch}, airplane wings and dog head in Fig.~\ref{fig:colorcnn+_visualization}) and \textit{textures} (\eg, cat tabby and face, vehicle wheels  and windshield, and  bird beak and breast in Fig.~\ref{fig:1-bit-batch} and Fig.~\ref{fig:colorcnn+_visualization}) are critical for single-label classifiers to correctly identify the image. 
Such findings also well align with some previous works \cite{geirhos2018imagenet,Goodfellow-et-al-2016} that both shapes and textures help neural network recognition. 



\textbf{Visualization results in \textit{large} color spaces.} 
However, as the color space size increases, pixels naturally support more structures (\eg, Fig.~\ref{fig:colorcnn+_visualization} (d), MedianCut results under 5-bit color space). 
Prioritizing key structures, visual fidelity of the ColorCNN quantized images  (\eg, Fig.~\ref{fig:colorcnn+_visualization} (e)-(g)) does not increase at a similar pace as the traditional methods when color space sizes increase. 
On the other hand, as shown in Fig.~\ref{fig:colorcnn+_visualization} (j), ColorCNN+ quantization results under 5-bit color spaces are significantly more visually accurate compared to ColorCNN outputs, and comparable to the traditional methods. 
With that said, ColorCNN+ results are still less visually accurate (\eg, dog before the blue sky) as it preserves not only visual fidelity under large color spaces but also key structures under small color spaces. 

\subsubsection{Recognition Accuracy}
\label{secsecsec:accuracy}

\textbf{Recognition accuracy in \textit{small} color spaces.} 
As shown in Fig.~\ref{fig:accuracy_curves}, for 1-bit and 2-bit color spaces, both ColorCNN and ColorCNN+ achieve significantly better performance on different datasets using different networks. 
These results verify that the structures identified and preserved by ColorCNNs are indeed critical for neural network recognition. 

\textbf{Recognition accuracy in \textit{large} color spaces.}
Under 5-bit and 6-bit color spaces, ColorCNN cannot compete with traditional color quantization methods in terms of network recognition accuracy. 
In fact, richer colors naturally support more structures, making 
the less visually accurate results from ColorCNN (Fig.~\ref{fig:colorcnn+_visualization} (g)) less competitive. 
In contrast, higher visual fidelity from ColorCNN+ results enables much easier recognition under \textit{large} color spaces. 
ColorCNN+ not only greatly outperforms vanilla ColorCNN, but also is competitive to the traditional color quantization methods like MedianCut, which we used as the ground-truth in our imitation cluster learning. 


\begin{figure}[t]
    \centering
    \begin{subfigure}[b]{\linewidth}
    \centering
        \includegraphics[width=\textwidth]{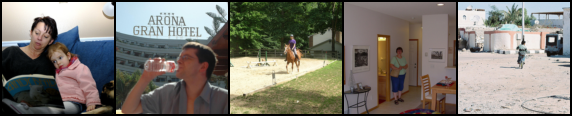}
        \caption{Original}
        \vspace{2mm}
    \end{subfigure}
    \begin{subfigure}[b]{\linewidth}
    \centering
        \includegraphics[width=\textwidth]{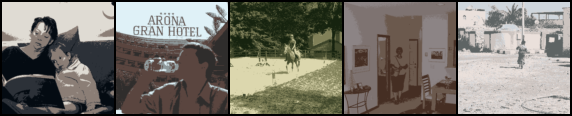}
        \caption{MedianCut}
        \vspace{2mm}
    \end{subfigure}
    \begin{subfigure}[b]{\linewidth}
    \centering
        \includegraphics[width=\textwidth]{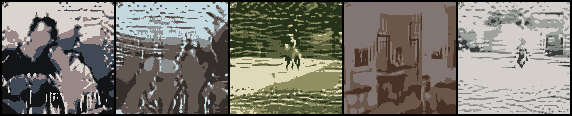}
        \caption{ColorCNN+}
    \end{subfigure}
\caption{Color quantization results (2-bit) for multi-label classifier on Pascal VOC 2012 dataset. 
}
\label{fig:voc}
\end{figure}

\begin{table}[]
\caption{Multi-label classification accuracy (\%) and mean average precision (mAP, \%) on Pascal VOC 2012 dataset using ResNet18 multi-label classifier. On original images, the classifier achieves 30.5\% accuracy and 55.3\% mAP. }
\label{tab:voc}
\resizebox{\linewidth}{!}{
\begin{tabular}{c|c|c|c|c|c|c|c}
\toprule
\multicolumn{2}{c|}{color space size   (bits)} & 1    & 2    & 3    & 4    & 5    & 6    \\ \hline
\multirow{2}{*}{accuracy} & MedianCut & 9.6  & 18.0 & 23.4 & 26.6 & 28.3 & 29.1 \\ \cline{2-8} 
                          & ColorCNN+ & \textbf{13.1} & \textbf{21.5} & \textbf{25.8} & \textbf{29.1} & \textbf{30.2} & \textbf{31.0} \\ \hline
\multirow{2}{*}{mAP}      & MedianCut & 24.5 & 39.6 & \textbf{46.7} & \textbf{50.4} & \textbf{52.4} & \textbf{53.5} \\ \cline{2-8} 
                          & ColorCNN+ & \textbf{33.4} & \textbf{41.1} & 44.9 & 49.7 & 51.9 & 53.2 \\ \bottomrule
\end{tabular}
}
\end{table}

\subsection{Discussion}
\label{secsec:discussion}

\subsubsection{Images with Multiple Salient Objects}

\textbf{Quantitatively} speaking (see Table~\ref{tab:voc}), ColorCNN+ shows its effectiveness in terms of multi-label classification accuracy (predicting all labels correctly), outperforming MedianCut on most settings. In terms of mean average precision (mAP), which is evaluated based on precision and recall at different sigmoid thresholds for each individual label, ColorCNN+ remains competitive, but falls slightly behind MedianCut at higher bit rates. In fact, as the multi-label classification loss pushes the ColorCNN+ output towards being more confident, evaluation at multiple thresholds (\eg, mAP) can be more demanding than evaluation at the default sigmoid threshold of 0.5 (\eg, multi-label classification accuracy).

\begin{figure}
\centering
    \begin{subfigure}[b]{0.19\linewidth}
    \centering
        \includegraphics[width=\textwidth]{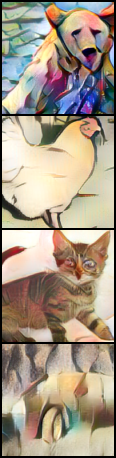}
        \caption{Original}
    \end{subfigure}
    \begin{subfigure}[b]{0.19\linewidth}
    \centering
        \includegraphics[width=\textwidth]{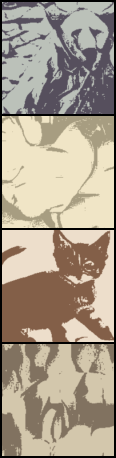}
        \caption{1-bit}
    \end{subfigure}
    \begin{subfigure}[b]{0.19\linewidth}
    \centering
        \includegraphics[width=\textwidth]{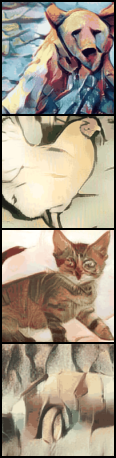}
        \caption{5-bit}
    \end{subfigure}
    \begin{subfigure}[b]{0.19\linewidth}
    \centering
        \includegraphics[width=\textwidth]{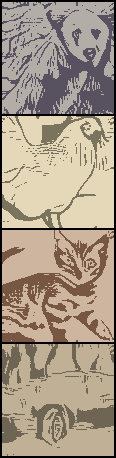}
        \caption{1-bit}
    \end{subfigure}
    \begin{subfigure}[b]{0.19\linewidth}
    \centering
        \includegraphics[width=\textwidth]{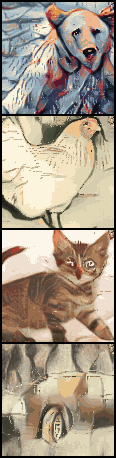}
        \caption{5-bit}
    \end{subfigure}
\caption{Color quantization results on stylized images. (a) is the original image. (b) and (c) are MedianCut results. (d) and (e) are ColorCNN+ results.
}
\label{fig:style}
\end{figure}

\begin{table}[]
\caption{Classification accuracy (\%) on Style14 dataset using ResNet18 classifier. On original (stylized but not color quantized) images, the classifier achieves 72.0\% accuracy. }
\label{tab:style}
\centering
\begin{tabular}{c|c|c|c|c|c|c}
\toprule
color space size (bits) & 1    & 2    & 3    & 4    & 5    & 6    \\ \hline
MedianCut        & 36.4 & 59.3 & 66.9 & 67.3 & 68.1 & 70.6 \\ \hline
ColorCNN+        & \textbf{66.4} & \textbf{69.7} & \textbf{71.0} & \textbf{70.9} & \textbf{71.8} & \textbf{71.7} \\ \bottomrule
\end{tabular}
\end{table}

\textbf{Qualitatively} speaking (see Fig.~\ref{fig:voc}), ColorCNN+ shows that for more complicated scenes (multiple objects-of-interest of different shapes and sizes, e.g., person, horse, motorbike, desk, bottle), the multi-label classifier prioritizes the coarse-grained overall shapes. In comparison, for single-label classification, as the classifiers only need to identify one object-of-interest, they additionally use fine-grained details like textures to further boost their performance. 
The multi-label setup is more demanding since the classifiers have to make out different objects of various shapes and sizes. Thus, the multi-label classifiers tend to focus more on the object shapes and outlines, which are more evident  and easier to identify. Fine-grained details, on the other hand, are still very important. However, due to limited capacity, the classifier oftentimes cannot take full advantage of such details to further boost the performance, as reflected in the ColorCNN+ outputs.

\subsubsection{Key Structures for Stylized Images}
As shown in Fig.~\ref{fig:style}, compared to traditional methods like MedianCut, we find ColorCNN+ better preserves both shapes (\eg, car body, chicken head) textures (\eg, the bear fur, chicken feather, cat tabby, and car wheels). Such visual differences also translate into network recognition accuracy improvements. In Table~\ref{tab:style}, under 1-bit color spaces, ColorCNN+ outperform MedianCut by +30.0\% accuracy, verifying its effectiveness. 

We point out that the stylized images still have textures. Though overshadowed by the added artistic style, they are still present. Neural network classifiers learn to exploit these fine-grained structures for better performance. And thus, they are present in ColorCNN+ outputs. 

\begin{table}[t]
\caption{Variant study for ColorCNN+ with ResNet18 classifier on CIFAR100 dataset. 
}
\label{tab:lambda}
\centering
\resizebox{\linewidth}{!}{
\begin{tabular}{c|c|c|c|c|c|c|c}
\toprule
\multicolumn{2}{c|}{color palette size (bits)} & 1     & 2     & 3     & 4     & 5     & 6     \\ \hline
\multicolumn{2}{c|}{MedianCut}                 & 10.2  & 26.1  & 44.0  & 57.8  & 66.2  & \textbf{70.8}  \\ \hline
\multicolumn{2}{c|}{ColorCNN}                  & 17.4  & 33.1  & 41.7  & 46.4  & 49.3  & 51.0  \\ \hline
\multirow{8}{*}{ColorCNN+} & $\lambda=0$           & \textbf{23.7}  & \textbf{45.7}  & 52.8  & 57.1  & 60.1  & 60.8  \\ \cline{2-8} 
                           & $\lambda=1$           & 23.0  & 44.5  & \textbf{53.1}  & 59.1  & 60.6  & 63.0  \\ \cline{2-8} 
                           & $\lambda=3$ (default) & 19.9  & 39.7  & 50.3  & \textbf{61.6}  & 65.6  & 67.9  \\ \cline{2-8} 
                           & $\lambda=10$          & 15.5  & 30.6  & 44.9  & 59.3  & \textbf{66.3}  & 70.4  \\ \cline{2-8} 
                           & w/o bottleneck      & 18.9 & 37.3 & 49.8 & 61.4 & 65.1 & 67.8 \\ \cline{2-8} 
                           & w/o top-$K$            & 17.9 & 39.0 & 49.8 & 61.0 & 65.7 & 68.1 \\ \cline{2-8} 
                           & w/o augmentation    & 14.9 & 32.7 & 47.1 & 59.7 & 65.4 & 68.3 \\ 
                          \cline{2-8} 
                          & CE$\rightarrow$KD    & 18.5 & 37.2 & 50.9 & 61.0 & 65.6 & 68.0 \\ 
                           \bottomrule
\end{tabular}
}
\end{table}

\subsubsection{Structure Preferences of Different Networks} 
Different classifiers have different preferences in terms of coarse-grained structures (\eg, object outlines and shapes) and fine-grained structures (\eg, object details and textures). ColorCNN+ faithfully reflects such structure preferences of different networks. 

First, for single-label classification (including that on both natural images and stylized images), classifiers learn to utilize \textit{both} coarse-grained structures (\eg, outlines of boat hull and cat body in Fig.~\ref{fig:1-bit-batch}, airplane wing and dog head in Fig.~\ref{fig:colorcnn+_visualization}, and animal bodies in Fig.~\ref{fig:style}) \textit{and} fine-grained structures (\eg, details of cat tabby and vehicle tire in Fig.~\ref{fig:1-bit-batch}, monkey fur and bird beak in Fig.~\ref{fig:colorcnn+_visualization}, and bear fur and chicken feather in Fig.~\ref{fig:style}). As classifiers only have to identify one object-of-interest in each image, they not only learn to identify the object shapes and outlines, but also take advantage of the informative details and textures to further improve their recognition accuracy. 

Second, for the same single-label classification dataset (see Fig.~\ref{fig:1-bit-batch}), we find a weaker classifier like AlexNet primarily focuses on the course-grained shapes and cannot take full advantage of fine-grained details due to its limited capacity (\eg, no details on the bird breast and the boat deck). On the other hand, stronger classifiers like VGG16 and ResNet18 attend more on the fine-grained details in addition to the coarse-grained shapes. 

Third, for multi-label classification, a task arguably more demanding than single-label classification, ColorCNN+ reflects rather different structural preferences in classifiers. As shown in Fig.~\ref{fig:voc}, coarse-grained structures like outlines of human, bottle, horse, desk, and motorbike are preserved. On the other hand, fine-grained details like human eyes and mouth, or background details like signs and trees, are usually left out. A possible reason for this is that the multi-label classification might turn out to be too demanding for the classifier. Although those details might help the recognition task, their limited capacity can prevent classifiers from taking full advantage of those details. 

Combing the aforementioned observations, we have the following findings. First, both coarse-grained and fine-grained structures help network recognition (coherent with previous findings in  \cite{geirhos2018imagenet,Goodfellow-et-al-2016}). Second, coarse-grained shapes and outlines are usually easier to learn and thus more commonly seen. Fine-grained details and textures, although helpful, can be difficult to learn for weak classifiers and demanding tasks due to limited network capacity.

\subsubsection{Key Structures: for Networks and for Humans}
In \textbf{small} color spaces, visualizations of ColorCNN+ results verify that both coarse-grained and fine-grained structures help neural network recognition. 
On the other hand, for human viewing, dithering generates a noise pattern to add details, but harms neural network recognition accuracy (Fig.~\ref{fig:accuracy_curves}). 
Comparing ColorCNNs (for network) and dithering (for human), we find the fine-grained structures like textures and details help both, but are rather distinct for their recognition. 

\begin{figure}
\centering
    \begin{subfigure}[b]{0.19\linewidth}
    \centering
        \includegraphics[width=\textwidth]{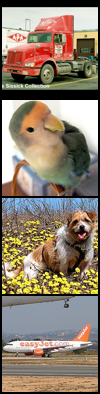}
        \caption{Original}
    \end{subfigure}
    \begin{subfigure}[b]{0.19\linewidth}
    \centering
        \includegraphics[width=\textwidth]{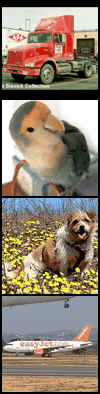}
        \caption{MCut}
    \end{subfigure}
    \begin{subfigure}[b]{0.19\linewidth}
    \centering
        \includegraphics[width=\textwidth]{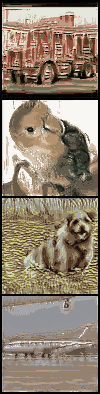}
        \caption{$\lambda=0$}
    \end{subfigure}
    \begin{subfigure}[b]{0.19\linewidth}
    \centering
        \includegraphics[width=\textwidth]{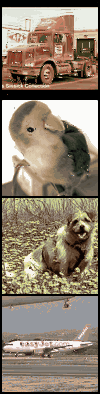}
        \caption{$\lambda=3$}
    \end{subfigure}
    \begin{subfigure}[b]{0.19\linewidth}
    \centering
        \includegraphics[width=\textwidth]{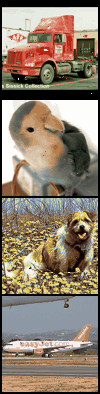}
        \caption{$\lambda=10$}
    \end{subfigure}
\caption{6-bit color quantization results. ``MCut'' in (a) is short for MedianCut. (b)-(e) are ColorCNN+ variants. $\lambda=3$ is the default hyper-parameter setting in ColorCNN+.
}
\label{fig:pixsim}
\end{figure}


\subsubsection{Key Structures vs. Accurate Colors}
In \textbf{large} color spaces, increasing the weight $\lambda$ for imitating traditional clustering-based methods effectively increases ColorCNN+ visual fidelity (Fig.~\ref{fig:pixsim}) and recognition accuracy (Table~\ref{tab:lambda}). 
This raises a question: \textit{is preserving visual fidelity (accurate colors) always beneficial to network recognition?}

\textit{The answer is No}. 
As shown in Fig.~\ref{fig:pixsim} and Table~\ref{tab:lambda}, increasing the weight $\lambda$ for the relationship preserving loss advocates visual fidelity, but does not always lead to higher recognition accuracy. In fact, for smaller color spaces (1 or 2 bits), increasing $\lambda$ actually decreases recognition accuracy. This aligns with our previous findings that ColorCNNs perform better than traditional color quantization method (\eg, MedianCut), even though traditional method results are still arguably more visually similar in terms of the pixel differences. It is also suggested that there exists a trade-off between these two concepts in small color spaces, and prioritizing the informative structures actually leads to higher neural network recognition accuracy.  
It is until larger color spaces (5 or 6 bits) that promoting the visual fidelity consistently improves the recognition accuracy, as the large color spaces naturally supports the structures. 
This justifies our choice of an intermediate weight for clustering loss $\lambda=3$ in all experiments.

\subsubsection{Low-bit Recognition Accuracy of Different Classifiers}
Color quantized images can achieve higher recognition accuracy with a weaker classifier. 
We compare the recognition accuracy with different classification networks in different rows of Fig.~\ref{fig:accuracy_curves}. 
It is found that a stronger classifier can have lower accuracy in an extremely small color space. 
For instance, on Tiny200 dataset, in 1-bit color space, AlexNet, VGG16, and ResNet18 recognize the MedianCut results with 11.3\%, 10.2\%, and 6.2\% accuracy, respectively. 
Stronger classifiers can apply stronger transformation to the image data, extracting more expressive features, thus having higher accuracy for full-color images. 
However, when the color space is limited, stronger transformation can lead to larger drifts in feature space, leading to poor generalizability. 

\subsection{Applications}

\begin{figure}[t]
\centering
\includegraphics[width=0.85\linewidth]{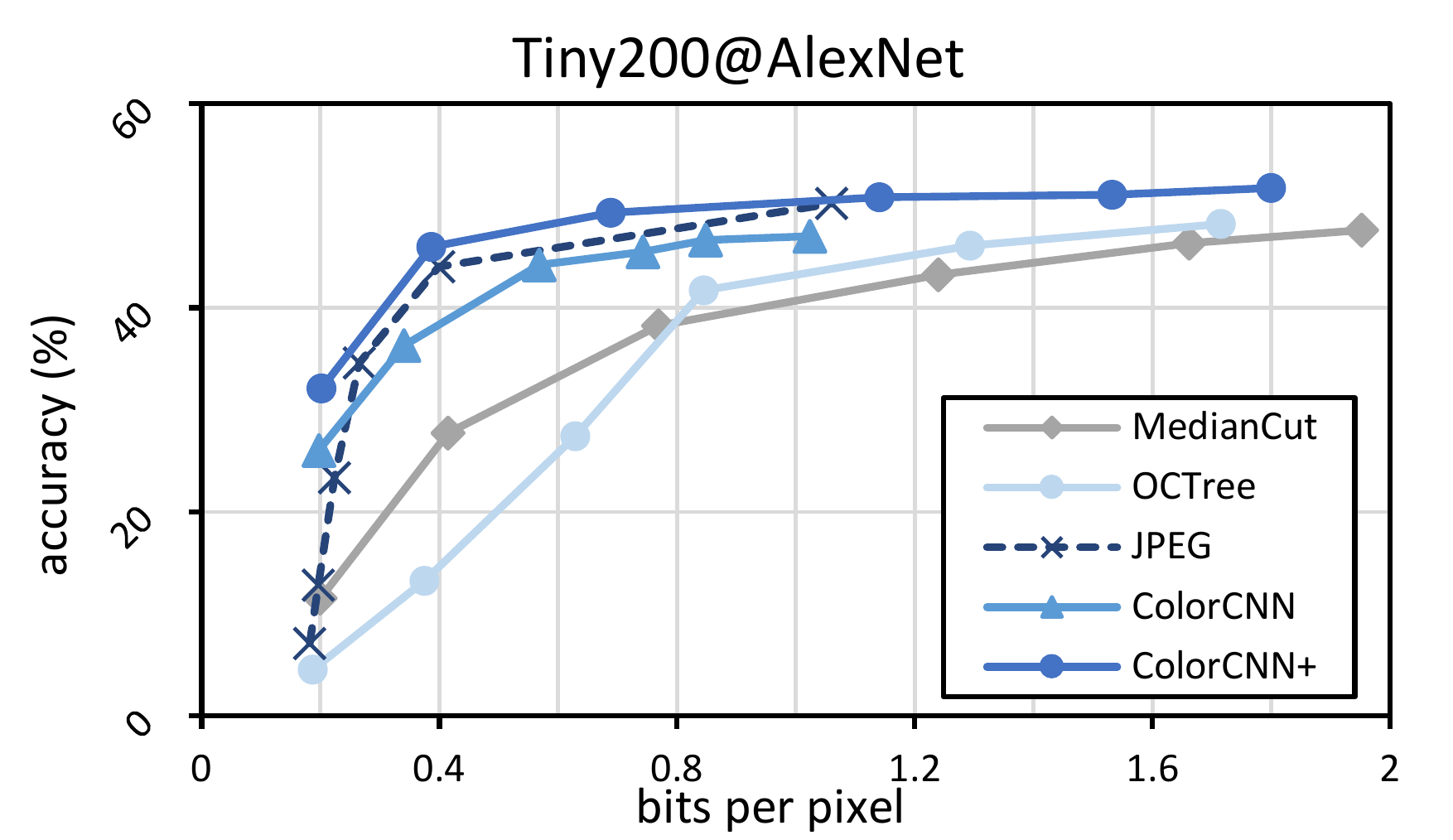}
\caption{Classification accuracy under different bitrate. Solid lines refer to color quantized image encoded via PNG. The dotted line refers to JPEG encoding as a reference. 
For color quantization methods, bitrate from low to high are quantized images under color space size from 1-bit to 6-bit. 
}
\label{fig:compression_curves}
\end{figure}

\textbf{ColorCNNs as image compression.} In Fig.~\ref{fig:compression_curves}, as the color space size grows from 1-bit to 6-bit, the quantized images take a higher bitrate when encoded with PNG, and are better recognized. 
When compared to traditional color quantization methods, ColorCNN can reach higher test accuracy under a lower bitrate. 
Moreover, under 0.2 bits per pixel, 1-bit ColorCNN quantization can even outperform JPEG compression by +13.2\%, which has arbitrary number of colors. 
ColorCNN+, on the other hand, constantly outperforms not only all other color quantization methods, but also JPEG image compression in terms of accuracy-to-compression ratio. This clearly demonstrates the effectiveness of ColorCNN+.
In fact, under 0.2 bits per pixel, 1-bit ColorCNN+ results outperform JPEG compression by +19.3\%.  
Moreover, ColorCNN+ supports multiple color space size configurations with a single model. Compared to vanilla ColorCNN, ColorCNN+ takes more bits to encode under larger color spaces, as the color spaces are more effectively utilized in ColorCNN+. 
Compared to other deep image compression methods that require a neural network decoder, the ColorCNNs with PNG encoding have minimal decoding computation requirement. 


\textbf{ColorCNNs as adversarial defense.}
Adversarial attacks add small perturbations to fool neural networks \cite{42503}. Lossy image compression like JPEG can effectively defend against such attacks \cite{das2018shield}. ColorCNNs, as lossy image compression methods, can also help defend against adversarial attacks. See appendix for experiments on adversarial defense.

\begin{figure}[t]
\centering
\includegraphics[width=\linewidth]{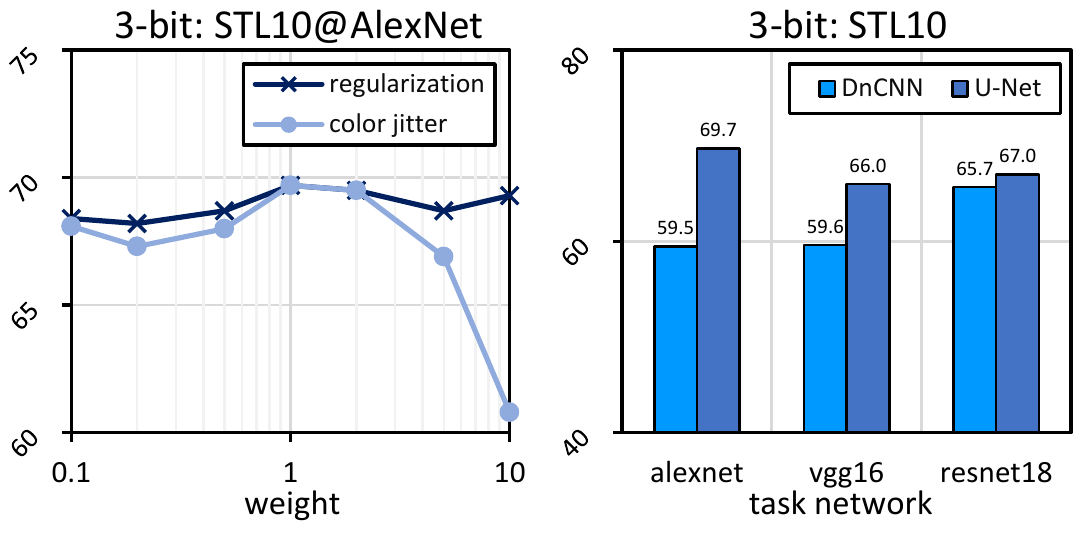}
\caption{ColorCNN performance (\%) comparison with different weights and different auto-encoder backbone. 
}
\label{fig:variants}
\end{figure}

\begin{table}[t]
\caption{ColorCNN and its variants under 3-bit color space, STL10 dataset, AlexNet classifier.}
\label{tab:ablation}
\resizebox{\linewidth}{!}{
\centering
\begin{tabular}{l|c|c|c}
\toprule
                   & \multicolumn{1}{l|}{Accuracy (\%)} & \multicolumn{1}{l|}{\#color/image} & \multicolumn{1}{l}{\#bit/pixel} \\ \hline
ColorCNN           & \textbf{69.7}                      & \textbf{8.0}                       & 0.425                               \\ \hline
w/o regularization & 67.5                               & 5.1                                & \textbf{0.323}                               \\ \hline
w/o color jitter   & 67.8                               & \textbf{8.0}                               & 0.390                               \\ \bottomrule
\end{tabular}
}
\end{table}

\subsection{Ablation and Variant Study}
\subsubsection{ColorCNN}

\textbf{Influence of color jitter.}
As shown in Table~\ref{tab:ablation}, without color jitter, the train-time quantization can be too easy for the pre-trained classifier. This can lead to overfitting, which further hurts accuracy by -1.9\%.
Too small or too large a color jitter can also result in a huge accuracy drop (Fig.~\ref{fig:variants}). This is because setting the weight too small or too large leads to either limited influence, or overshadowing anything else. 

\textbf{Influence of color maximum regularization.}
We find that removing the color maximum regularization term causes an accuracy drop in Table~\ref{tab:ablation}. In fact, without the regularization, fewer colors are chosen during test-time. This is because the softmax color filling during training can introduce more colors in the image, as shown in Fig.~\ref{fig:comparison}. 
When the regularization weight is too small or too high, ColorCNN performance decreases (Fig.~\ref{fig:variants}). 




\textbf{Influence of feature extractor backbone.}
As shown in Fig.~\ref{fig:variants}, when the auto-encoder backbone is replaced with DnCNN~\cite{zhang2017beyond}, the ColorCNN performance degrades under all classification networks. Unlike U-Net, DnCNN does not have bypasses to maintain the fine-grained structures. As a result, its quantization results might have structure misalignment, which hurts classification accuracy.


\subsubsection{ColorCNN+}

\textbf{Influence of the bottleneck layer.}
As in Table~\ref{tab:lambda}, the removal of the bottleneck layer results in accuracy drops in small and medium-sized color spaces (-2.0\%, -0.7\%, -0.5\%, and -0.6\% for 1-4 bit color spaces), and slight improvements in large color spaces (+0.1\% and +0.2\% for 5 and 6-bit color spaces). The performance drops confirms that redundant and irrelevant information can lead to overfitting under small color spaces. On the other hand, the performance increases under large color spaces align with our understanding that more difficult decisions benefit from more information. 
Overall, the performance differences under small and medium-sized color spaces largely outweigh those in large color spaces, proving the effectiveness of this low-dimension bottleneck. 

\textbf{Influence of $K$ non-zero terms.}
We find that removing the top-$K$ non-zero mask in the color probability map $m_C\left(\bm{x}\right)$ leads to consistent performance drops in Table~\ref{tab:lambda}. This verifies its effectiveness in combating overfitting, as it directly forces the train-time forward pass approximation to behave more like the test-time counterpart.

\textbf{Influence of the pixel-wise relationship preserving loss.}
In Fig.~\ref{fig:pixsim} and Table~\ref{tab:lambda}, we find that the removal of the imitation learning loss results in undesirable visual quality (limited usage of the color spaces) and inferior recognition accuracy under large color spaces, verifying its effectiveness. 
With that said, the inclusion of the pixel-wise relationship preserving loss also results in accuracy drops for small color spaces. In this regard, we refer readers to Section~\ref{secsec:discussion}, where we discuss the trade-off between accurate colors and key structure, especially in small color spaces.

\begin{figure}[t]
\centering
\includegraphics[width=0.98\linewidth]{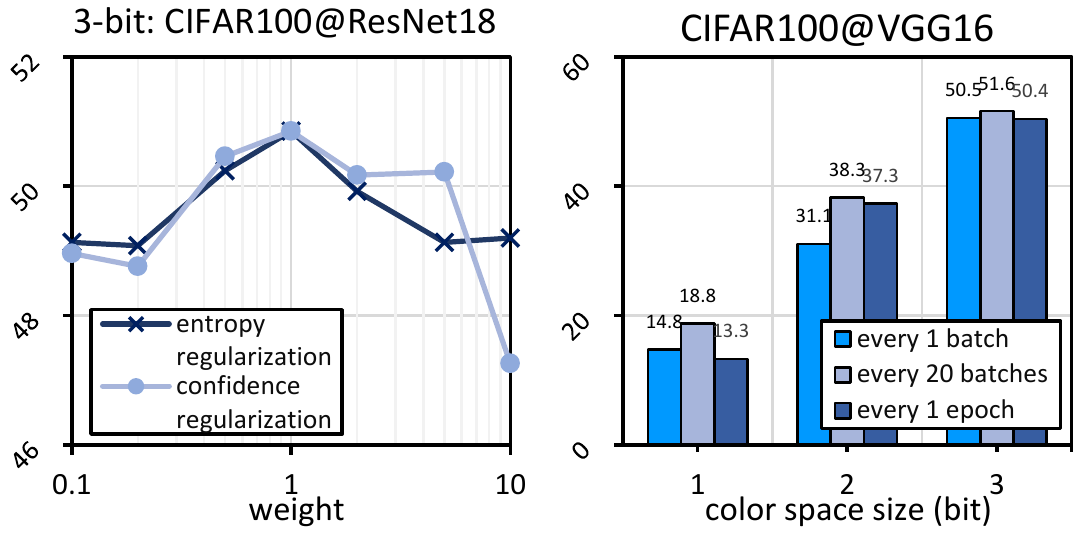}
\caption{\textbf{Left:} ColorCNN+ performance comparison with different weights for $R_\text{info}$ and $R_\text{conf}$. \textbf{Right:} Training ColorCNN+ using different frequency for task selection update. 
}
\label{fig:regularization_task_selection}
\end{figure}

\textbf{Information entropy and confidence regularization terms.}
In Fig.~\ref{fig:regularization_task_selection} left, we find that too small or too large values of the two regularization terms lead to worse performance. These results probably due to too little regularizations are not strong enough,
to combat the limited color choices (information entropy regularization) or overfitting (confidence regularization). On the other hand, 
while too much regularizations can overshadow other loss components like the cross-entropy loss or the pixel-wise relationship preserving loss. 
We settle at setting $\alpha=1$ and $\beta=1$ for all our experiments.

\textbf{Pace of task selection change.}
As ColorCNN+ natively supports multiple color space sizes in one model, during training, how to choose the color space size becomes an important question. This is referred as task selection (see Section~\ref{secsecsec:task_selection}). As shown in Fig.~\ref{fig:regularization_task_selection}, we find that neither too fast (\ie, every batch) nor slow (\ie, every epoch) the pace in task change gives better results than the proposed intermediate pace for changing the task (color space size) in training. This aligns with the previous study \cite{wu2020multigrid} on task selection for video understanding.

\textbf{Augmentation on the quantized images.}
As shown in Table~\ref{tab:lambda}, the absence of post-time augmentations on quantized images results in performance drops except for 6-bit color spaces. Especially, for small color spaces (1 and 2-bit), the resulting variant is even outperformed by the vanilla ColorCNN. This verifies that the augmentation is vital for learning the informative structures in ColorCNN+, and that augmentations on the quantized images effectively prevent overfitting in ColorCNN+ training.

\textbf{Knowledge distillation instead of cross-entropy loss.}
In the absence of ground-truth labels for the images, we can replace the cross-entropy loss in Eq.~\ref{eq:combine_loss} with knowledge distillation loss \cite{hinton2015distilling}, which uses the classifier output on the original image as soft targets. As shown in Table~\ref{tab:lambda}, using knowledge distillation loss instead of cross-entropy loss achieves similar performance. 

\section{Conclusion}\label{sec:conclusion}

In this paper, we investigate the scientific problem of keeping informative structures with limited colors. Specifically, it is found that using the task network and a specifically designed architecture, ColorCNN, we can preserve the informative structures to enable neural network recognition in small color spaces. By imitating traditional clustering-based color quantization methods, ColorCNN+, an updated architecture that supports multiple color space size configurations can output visually accurate results under large color spaces. Extensive experiments showcase the identified informative structures and enable study on when and how the informative structures and accurate colors help neural network recognition.

\ifCLASSOPTIONcaptionsoff
  \newpage
\fi

\bibliographystyle{IEEEtran}
\bibliography{main}

\begin{thebibliography}{10}
\providecommand{\url}[1]{#1}
\csname url@samestyle\endcsname
\providecommand{\newblock}{\relax}
\providecommand{\bibinfo}[2]{#2}
\providecommand{\BIBentrySTDinterwordspacing}{\spaceskip=0pt\relax}
\providecommand{\BIBentryALTinterwordstretchfactor}{4}
\providecommand{\BIBentryALTinterwordspacing}{\spaceskip=\fontdimen2\font plus
\BIBentryALTinterwordstretchfactor\fontdimen3\font minus
  \fontdimen4\font\relax}
\providecommand{\BIBforeignlanguage}[2]{{%
\expandafter\ifx\csname l@#1\endcsname\relax
\typeout{** WARNING: IEEEtran.bst: No hyphenation pattern has been}%
\typeout{** loaded for the language `#1'. Using the pattern for}%
\typeout{** the default language instead.}%
\else
\language=\csname l@#1\endcsname
\fi
#2}}
\providecommand{\BIBdecl}{\relax}
\BIBdecl

\bibitem{heckbert1982color}
P.~Heckbert, \emph{Color image quantization for frame buffer display}.\hskip
  1em plus 0.5em minus 0.4em\relax ACM, 1982, vol.~16, no.~3.

\bibitem{zhou2016learning}
B.~Zhou, A.~Khosla, A.~Lapedriza, A.~Oliva, and A.~Torralba, ``Learning deep
  features for discriminative localization,'' in \emph{Proceedings of the IEEE
  conference on computer vision and pattern recognition}, 2016, pp. 2921--2929.

\bibitem{orchard1991color}
M.~T. Orchard and C.~A. Bouman, ``Color quantization of images,'' \emph{IEEE
  transactions on signal processing}, vol.~39, no.~12, pp. 2677--2690, 1991.

\bibitem{floyd1976adaptive}
R.~Floyd and L.~Steinberg, ``An adaptive technique for spatial grayscale,'' in
  \emph{Proceedings of the Society of Information Display}, vol.~17, 1976, pp.
  78--84.

\bibitem{gervautz1988simple}
M.~Gervautz and W.~Purgathofer, ``A simple method for color quantization:
  Octree quantization,'' in \emph{New trends in computer graphics}.\hskip 1em
  plus 0.5em minus 0.4em\relax Springer, 1988, pp. 219--231.

\bibitem{hou2020learning}
Y.~Hou, L.~Zheng, and S.~Gould, ``Learning to structure an image with few
  colors,'' in \emph{Proceedings of the IEEE/CVF Conference on Computer Vision
  and Pattern Recognition}, 2020, pp. 10\,116--10\,125.

\bibitem{deng1999peer}
Y.~Deng, C.~Kenney, M.~S. Moore, and B.~Manjunath, ``Peer group filtering and
  perceptual color image quantization,'' in \emph{ISCAS'99. Proceedings of the
  1999 IEEE International Symposium on Circuits and Systems VLSI (Cat. No.
  99CH36349)}, vol.~4.\hskip 1em plus 0.5em minus 0.4em\relax IEEE, 1999, pp.
  21--24.

\bibitem{achanta2012slic}
R.~Achanta, A.~Shaji, K.~Smith, A.~Lucchi, P.~Fua, and S.~S{\"u}sstrunk, ``Slic
  superpixels compared to state-of-the-art superpixel methods,'' \emph{IEEE
  transactions on pattern analysis and machine intelligence}, vol.~34, no.~11,
  pp. 2274--2282, 2012.

\bibitem{deng2001unsupervised}
Y.~Deng and B.~Manjunath, ``Unsupervised segmentation of color-texture regions
  in images and video,'' \emph{IEEE transactions on pattern analysis and
  machine intelligence}, vol.~23, no.~8, pp. 800--810, 2001.

\bibitem{wu1992color}
X.~Wu, ``Color quantization by dynamic programming and principal analysis,''
  \emph{ACM Transactions on Graphics (TOG)}, vol.~11, no.~4, pp. 348--372,
  1992.

\bibitem{boutell1997png}
T.~Boutell and T.~Lane, ``Png (portable network graphics) specification version
  1.0,'' \emph{Network Working Group}, pp. 1--102, 1997.

\bibitem{wallace1992jpeg}
G.~K. Wallace, ``The jpeg still picture compression standard,'' \emph{IEEE
  transactions on consumer electronics}, vol.~38, no.~1, pp. xviii--xxxiv,
  1992.

\bibitem{skodras2001jpeg}
A.~Skodras, C.~Christopoulos, and T.~Ebrahimi, ``The jpeg 2000 still image
  compression standard,'' \emph{IEEE Signal processing magazine}, vol.~18,
  no.~5, pp. 36--58, 2001.

\bibitem{poynton2012digital}
C.~Poynton, \emph{Digital video and HD: Algorithms and Interfaces}.\hskip 1em
  plus 0.5em minus 0.4em\relax Elsevier, 2012.

\bibitem{oord2016pixel}
A.~v.~d. Oord, N.~Kalchbrenner, and K.~Kavukcuoglu, ``Pixel recurrent neural
  networks,'' \emph{arXiv preprint arXiv:1601.06759}, 2016.

\bibitem{johnston2018improved}
N.~Johnston, D.~Vincent, D.~Minnen, M.~Covell, S.~Singh, T.~Chinen,
  S.~Jin~Hwang, J.~Shor, and G.~Toderici, ``Improved lossy image compression
  with priming and spatially adaptive bit rates for recurrent networks,'' in
  \emph{Proceedings of the IEEE Conference on Computer Vision and Pattern
  Recognition}, 2018, pp. 4385--4393.

\bibitem{toderici2017full}
G.~Toderici, D.~Vincent, N.~Johnston, S.~Jin~Hwang, D.~Minnen, J.~Shor, and
  M.~Covell, ``Full resolution image compression with recurrent neural
  networks,'' in \emph{Proceedings of the IEEE Conference on Computer Vision
  and Pattern Recognition}, 2017, pp. 5306--5314.

\bibitem{mentzer2019practical}
F.~Mentzer, E.~Agustsson, M.~Tschannen, R.~Timofte, and L.~V. Gool, ``Practical
  full resolution learned lossless image compression,'' in \emph{Proceedings of
  the IEEE Conference on Computer Vision and Pattern Recognition}, 2019, pp.
  10\,629--10\,638.

\bibitem{li2018learning}
M.~Li, W.~Zuo, S.~Gu, D.~Zhao, and D.~Zhang, ``Learning convolutional networks
  for content-weighted image compression,'' in \emph{Proceedings of the IEEE
  Conference on Computer Vision and Pattern Recognition}, 2018, pp. 3214--3223.

\bibitem{van2016conditional}
A.~Van~den Oord, N.~Kalchbrenner, L.~Espeholt, O.~Vinyals, A.~Graves
  \emph{et~al.}, ``Conditional image generation with pixelcnn decoders,'' in
  \emph{Advances in neural information processing systems}, 2016, pp.
  4790--4798.

\bibitem{agustsson2018generative}
E.~Agustsson, M.~Tschannen, F.~Mentzer, R.~Timofte, and L.~Van~Gool,
  ``Generative adversarial networks for extreme learned image compression,''
  \emph{arXiv preprint arXiv:1804.02958}, 2018.

\bibitem{balle2016end}
J.~Ball{\'e}, V.~Laparra, and E.~P. Simoncelli, ``End-to-end optimized image
  compression,'' \emph{arXiv preprint arXiv:1611.01704}, 2016.

\bibitem{balle2018variational}
J.~Ball{\'e}, D.~Minnen, S.~Singh, S.~J. Hwang, and N.~Johnston, ``Variational
  image compression with a scale hyperprior,'' \emph{arXiv preprint
  arXiv:1802.01436}, 2018.

\bibitem{liu2019machine}
Z.~Liu, X.~Xu, T.~Liu, Q.~Liu, Y.~Wang, Y.~Shi, W.~Wen, M.~Huang, H.~Yuan, and
  J.~Zhuang, ``Machine vision guided 3d medical image compression for efficient
  transmission and accurate segmentation in the clouds,'' \emph{arXiv preprint
  arXiv:1904.08487}, 2019.

\bibitem{wei2019learning}
X.~Wei, I.~A. Barsan, S.~Wang, J.~Martinez, and R.~Urtasun, ``Learning to
  localize through compressed binary maps,'' in \emph{Proceedings of the IEEE
  Conference on Computer Vision and Pattern Recognition}, 2019, pp.
  10\,316--10\,324.

\bibitem{camposeco2019hybrid}
F.~Camposeco, A.~Cohen, M.~Pollefeys, and T.~Sattler, ``Hybrid scene
  compression for visual localization,'' in \emph{Proceedings of the IEEE
  Conference on Computer Vision and Pattern Recognition}, 2019, pp. 7653--7662.

\bibitem{kohonen1982self}
T.~Kohonen, ``Self-organized formation of topologically correct feature maps,''
  \emph{Biological cybernetics}, vol.~43, no.~1, pp. 59--69, 1982.

\bibitem{yang2017towards}
B.~Yang, X.~Fu, N.~D. Sidiropoulos, and M.~Hong, ``Towards k-means-friendly
  spaces: Simultaneous deep learning and clustering,'' in \emph{international
  conference on machine learning}.\hskip 1em plus 0.5em minus 0.4em\relax PMLR,
  2017, pp. 3861--3870.

\bibitem{xie2016unsupervised}
J.~Xie, R.~Girshick, and A.~Farhadi, ``Unsupervised deep embedding for
  clustering analysis,'' in \emph{International conference on machine
  learning}, 2016, pp. 478--487.

\bibitem{li2018discriminatively}
F.~Li, H.~Qiao, and B.~Zhang, ``Discriminatively boosted image clustering with
  fully convolutional auto-encoders,'' \emph{Pattern Recognition}, vol.~83, pp.
  161--173, 2018.

\bibitem{aljalbout2018clustering}
E.~Aljalbout, V.~Golkov, Y.~Siddiqui, M.~Strobel, and D.~Cremers, ``Clustering
  with deep learning: Taxonomy and new methods,'' \emph{arXiv preprint
  arXiv:1801.07648}, 2018.

\bibitem{hsu2017cnn}
C.-C. Hsu and C.-W. Lin, ``Cnn-based joint clustering and representation
  learning with feature drift compensation for large-scale image data,''
  \emph{IEEE Transactions on Multimedia}, vol.~20, no.~2, pp. 421--429, 2017.

\bibitem{wang2016learning}
Z.~Wang, S.~Chang, J.~Zhou, M.~Wang, and T.~S. Huang, ``Learning a
  task-specific deep architecture for clustering,'' in \emph{Proceedings of the
  2016 SIAM International Conference on Data Mining}.\hskip 1em plus 0.5em
  minus 0.4em\relax SIAM, 2016, pp. 369--377.

\bibitem{yang2016joint}
J.~Yang, D.~Parikh, and D.~Batra, ``Joint unsupervised learning of deep
  representations and image clusters,'' in \emph{Proceedings of the IEEE
  conference on computer vision and pattern recognition}, 2016, pp. 5147--5156.

\bibitem{chen2017unsupervised}
D.~Chen, J.~Lv, and Y.~Zhang, ``Unsupervised multi-manifold clustering by
  learning deep representation,'' in \emph{Workshops at the thirty-first AAAI
  conference on artificial intelligence}, 2017.

\bibitem{ghasedi2017deep}
K.~Ghasedi~Dizaji, A.~Herandi, C.~Deng, W.~Cai, and H.~Huang, ``Deep clustering
  via joint convolutional autoencoder embedding and relative entropy
  minimization,'' in \emph{Proceedings of the IEEE international conference on
  computer vision}, 2017, pp. 5736--5745.

\bibitem{hu2017learning}
W.~Hu, T.~Miyato, S.~Tokui, E.~Matsumoto, and M.~Sugiyama, ``Learning discrete
  representations via information maximizing self-augmented training,'' in
  \emph{International conference on machine learning}.\hskip 1em plus 0.5em
  minus 0.4em\relax PMLR, 2017, pp. 1558--1567.

\bibitem{ronneberger2015u}
O.~Ronneberger, P.~Fischer, and T.~Brox, ``U-net: Convolutional networks for
  biomedical image segmentation,'' in \emph{International Conference on Medical
  image computing and computer-assisted intervention}.\hskip 1em plus 0.5em
  minus 0.4em\relax Springer, 2015, pp. 234--241.

\bibitem{huang2014deep}
P.~Huang, Y.~Huang, W.~Wang, and L.~Wang, ``Deep embedding network for
  clustering,'' in \emph{2014 22nd International conference on pattern
  recognition}.\hskip 1em plus 0.5em minus 0.4em\relax IEEE, 2014, pp.
  1532--1537.

\bibitem{tung2019similarity}
F.~Tung and G.~Mori, ``Similarity-preserving knowledge distillation,'' in
  \emph{Proceedings of the IEEE International Conference on Computer Vision},
  2019, pp. 1365--1374.

\bibitem{li2020semantic}
Z.~Li, R.~Jiang, and P.~Aarabi, ``Semantic relation preserving knowledge
  distillation for image-to-image translation,'' in \emph{European conference
  on computer vision}.\hskip 1em plus 0.5em minus 0.4em\relax Springer, 2020.

\bibitem{hou2021visualizing}
Y.~Hou and L.~Zheng, ``Visualizing adapted knowledge in domain transfer,'' in
  \emph{Proceedings of the IEEE/CVF Conference on Computer Vision and Pattern
  Recognition}, 2021, pp. 13\,824--13\,833.

\bibitem{shannon1948mathematical}
C.~E. Shannon, ``A mathematical theory of communication,'' \emph{Bell system
  technical journal}, vol.~27, no.~3, pp. 379--423, 1948.

\bibitem{zhong2020random}
Z.~Zhong, L.~Zheng, G.~Kang, S.~Li, and Y.~Yang, ``Random erasing data
  augmentation,'' in \emph{Proceedings of the AAAI Conference on Artificial
  Intelligence}, vol.~34, no.~07, 2020, pp. 13\,001--13\,008.

\bibitem{wu2020multigrid}
C.-Y. Wu, R.~Girshick, K.~He, C.~Feichtenhofer, and P.~Krahenbuhl, ``A
  multigrid method for efficiently training video models,'' in
  \emph{Proceedings of the IEEE/CVF Conference on Computer Vision and Pattern
  Recognition}, 2020, pp. 153--162.

\bibitem{krizhevsky2012imagenet}
A.~Krizhevsky, I.~Sutskever, and G.~E. Hinton, ``Imagenet classification with
  deep convolutional neural networks,'' in \emph{Advances in neural information
  processing systems}, 2012, pp. 1097--1105.

\bibitem{simonyan2014very}
K.~Simonyan and A.~Zisserman, ``Very deep convolutional networks for
  large-scale image recognition,'' \emph{arXiv preprint arXiv:1409.1556}, 2014.

\bibitem{he2016deep}
K.~He, X.~Zhang, S.~Ren, and J.~Sun, ``Deep residual learning for image
  recognition,'' in \emph{Proceedings of the IEEE conference on computer vision
  and pattern recognition}, 2016, pp. 770--778.

\bibitem{krizhevsky2009learning}
A.~Krizhevsky, G.~Hinton \emph{et~al.}, ``Learning multiple layers of features
  from tiny images,'' Citeseer, Tech. Rep., 2009.

\bibitem{coates2011analysis}
A.~Coates, A.~Ng, and H.~Lee, ``An analysis of single-layer networks in
  unsupervised feature learning,'' in \emph{Proceedings of the fourteenth
  international conference on artificial intelligence and statistics}, 2011,
  pp. 215--223.

\bibitem{le2015tiny}
Y.~Le and X.~Yang, ``Tiny imagenet visual recognition challenge,'' \emph{CS
  231N}, 2015.

\bibitem{deng2009imagenet}
J.~Deng, W.~Dong, R.~Socher, L.-J. Li, K.~Li, and L.~Fei-Fei, ``Imagenet: A
  large-scale hierarchical image database,'' in \emph{2009 IEEE conference on
  computer vision and pattern recognition}.\hskip 1em plus 0.5em minus
  0.4em\relax Ieee, 2009, pp. 248--255.

\bibitem{everingham2010pascal}
M.~Everingham, L.~Van~Gool, C.~K. Williams, J.~Winn, and A.~Zisserman, ``The
  pascal visual object classes (voc) challenge,'' \emph{International journal
  of computer vision}, vol.~88, no.~2, pp. 303--338, 2010.

\bibitem{geirhos2018imagenet}
R.~Geirhos, P.~Rubisch, C.~Michaelis, M.~Bethge, F.~A. Wichmann, and
  W.~Brendel, ``Imagenet-trained cnns are biased towards texture; increasing
  shape bias improves accuracy and robustness,'' \emph{arXiv preprint
  arXiv:1811.12231}, 2018.

\bibitem{smith2019super}
L.~N. Smith and N.~Topin, ``Super-convergence: Very fast training of neural
  networks using large learning rates,'' in \emph{Artificial Intelligence and
  Machine Learning for Multi-Domain Operations Applications}, vol. 11006.\hskip
  1em plus 0.5em minus 0.4em\relax International Society for Optics and
  Photonics, 2019, p. 1100612.

\bibitem{loshchilov2016sgdr}
I.~Loshchilov and F.~Hutter, ``Sgdr: Stochastic gradient descent with warm
  restarts,'' \emph{arXiv preprint arXiv:1608.03983}, 2016.

\bibitem{Goodfellow-et-al-2016}
I.~Goodfellow, Y.~Bengio, and A.~Courville, \emph{Deep Learning}.\hskip 1em
  plus 0.5em minus 0.4em\relax MIT Press, 2016,
  \url{http://www.deeplearningbook.org}.

\bibitem{42503}
\BIBentryALTinterwordspacing
C.~Szegedy, W.~Zaremba, I.~Sutskever, J.~Bruna, D.~Erhan, I.~Goodfellow, and
  R.~Fergus, ``Intriguing properties of neural networks,'' in
  \emph{International Conference on Learning Representations}, 2014. [Online].
  Available: \url{http://arxiv.org/abs/1312.6199}
\BIBentrySTDinterwordspacing

\bibitem{das2018shield}
N.~Das, M.~Shanbhogue, S.-T. Chen, F.~Hohman, S.~Li, L.~Chen, M.~E. Kounavis,
  and D.~H. Chau, ``Shield: Fast, practical defense and vaccination for deep
  learning using jpeg compression,'' in \emph{Proceedings of the 24th ACM
  SIGKDD International Conference on Knowledge Discovery \& Data Mining}, 2018,
  pp. 196--204.

\bibitem{zhang2017beyond}
K.~Zhang, W.~Zuo, Y.~Chen, D.~Meng, and L.~Zhang, ``Beyond a gaussian denoiser:
  Residual learning of deep cnn for image denoising,'' \emph{IEEE Transactions
  on Image Processing}, vol.~26, no.~7, pp. 3142--3155, 2017.

\bibitem{hinton2015distilling}
G.~Hinton, O.~Vinyals, and J.~Dean, ``Distilling the knowledge in a neural
  network,'' \emph{arXiv preprint arXiv:1503.02531}, 2015.

\end{thebibliography}

%

\newpage


\begin{IEEEbiography}[{\includegraphics[width=1in,height=1.25in,clip,keepaspectratio]{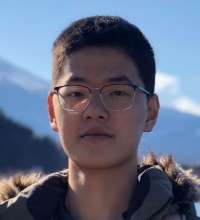}}]{Yunzhong Hou}
received his bachelor degree in electronic engineering from Tsinghua University in 2018. He is now working towards a PhD degree at Australian National University under the supervision of Dr. Liang Zheng and Prof. Stephen Gould. His research interests lies in computer vision and deep learning. 
\end{IEEEbiography}


\begin{IEEEbiography}[{\includegraphics[width=1in,height=1.25in,clip,keepaspectratio]{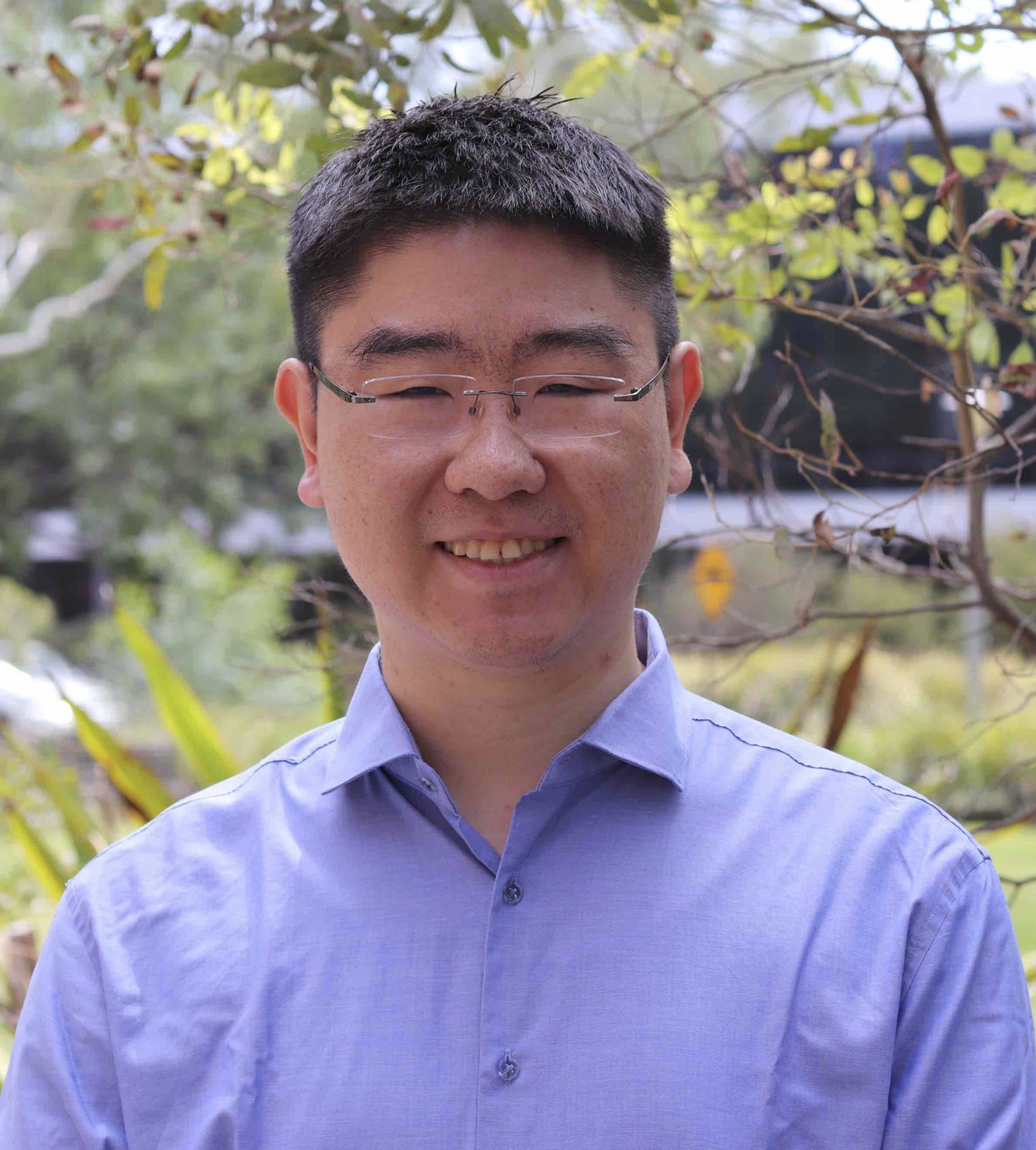}}]{Liang Zheng}
is a Lecturer and a Computer Science Futures Fellow in the Research School of Computer Science, Australian National University. He received the PhD degree in Electronic Engineering from Tsinghua University, China, in 2015, and the B.E. degree in Life Science from Tsinghua University, China, in 2010. He was a postdoc researcher in the Center for Artificial Intelligence, University of Technology Sydney, Australia. His research interests include image retrieval, classification, and person re-identification.
\end{IEEEbiography}


\begin{IEEEbiography}[{\includegraphics[width=1in,height=1.25in,clip,keepaspectratio]{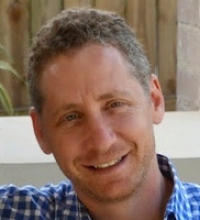}}]{Stephen Gould}
received the BSc degree in mathematics and computer science and BE degree in electrical engineering from the University of Sydney, in 1994 and 1996, respectively, the MS degree in electrical engineering from Stanford University, in 1998, and the PhD degree from Stanford University, in 2010. He is a professor with the Research School of Computer Science, College of Engineering and Computer Science, Australian National University. His research interests are in computer and robotic vision, machine learning, probabilistic graphical models, and optimization.
\end{IEEEbiography}


\vfill


\end{document}